\pdfoutput=1

\documentclass[11pt]{article}

\usepackage{ACL2023}

\usepackage{times}
\usepackage{latexsym}

\usepackage[T1]{fontenc}

\usepackage[utf8]{inputenc}

\usepackage{microtype}

\usepackage{inconsolata}

\usepackage{graphicx}
\usepackage{amsmath}

\title{B2T Connection: Serving Stability and Performance in Deep Transformers}

\author{Sho Takase{$^\dagger$}\thanks{\ \ A part of this work was done when the author was at Tokyo Institute of Technology.} \hspace{5mm}
  Shun Kiyono{$^\dagger$} \hspace{5mm}
  Sosuke Kobayashi{$^\ddagger$} \hspace{5mm}
  Jun Suzuki{$^\ddagger$} \\
  {$^\dagger$}LINE Corporation \hspace{25mm} {$^\ddagger$}Tohoku University \\
  \texttt{\{sho.takase, shun.kiyono\}@linecorp.com} \\
  \texttt{sosk@preferred.jp} \\
  \texttt{jun.suzuki@tohoku.ac.jp} \\
  }

\begin{document}
\maketitle
\begin{abstract}
From the perspective of the layer normalization (LN) positions, the architectures of Transformers can be categorized into two types: Post-LN and Pre-LN.
Recent Transformers tend to be Pre-LN because, in Post-LN with deep Transformers (e.g., those with ten or more layers), the training is often unstable, resulting in useless models.
However, Post-LN has consistently achieved better performance than Pre-LN in relatively shallow Transformers (e.g., those with six or fewer layers).
This study first investigates the reason for these discrepant observations empirically and theoretically and made the following discoveries: 1, the LN in Post-LN is the main source of the vanishing gradient problem that leads to unstable training, whereas Pre-LN prevents it, and 2, Post-LN tends to preserve larger gradient norms in higher layers during the back-propagation, which may lead to effective training.
Exploiting the new findings, we propose a method that can provide both high stability and effective training by a simple modification of Post-LN.
We conduct experiments on a wide range of text generation tasks.
The experimental results demonstrate that our method outperforms Pre-LN, and enables stable training regardless of the shallow or deep layer settings.
Our code is publicly available at \href{https://github.com/takase/b2t_connection}{https://github.com/takase/b2t\_connection}.
\end{abstract}

\section{Introduction}
\label{sec:intro}

To prevent the vanishing (or exploding) gradient problem in the training of a deep neural network (DNN), various techniques, such as batch normalization~\cite{pmlr-v37-ioffe15} and residual connection~\cite{NIPS2015_215a71a1,7780459}, have been proposed and widely used in almost all recent DNNs.
Transformer~\cite{NIPS2017_7181} employs the layer normalization~\cite{ba2016layer} for this purpose.
Transformer is currently the most successful model architecture in DNNs.
It was firstly developed for applying sequence-to-sequence tasks, such as machine translation~\cite{NIPS2017_7181}, summarization~\cite{takase-okazaki-2019-positional}, and automatic speech recognition (ASR)~\cite{wang-etal-2020-fairseq}, and is currently used in speech, vision, and many other information processing research fields.

\begin{figure}[!t]
  \centering 
    \includegraphics[width=8cm]{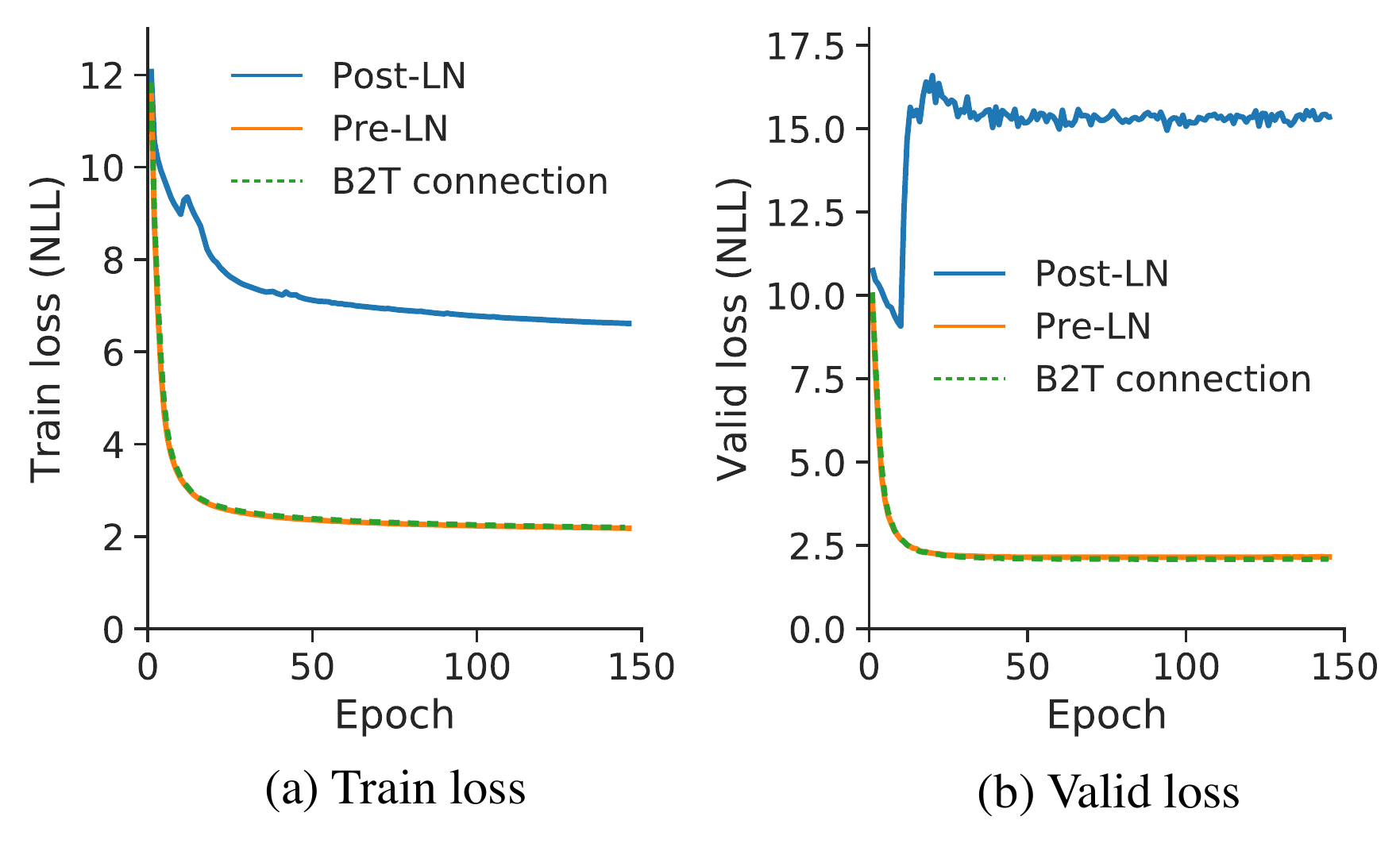}
   \caption{Loss values of 18 layered Transformer-based encoder-decoder architectures.}
   \label{fig:loss4encdec}
\end{figure}

As reported in the batch normalization literature~\cite{10.1007/978-3-319-46493-0_38}, the position of the normalization layers primarily affects both the stability and resultant performance of a trained model.
In Transformers, some previous studies have investigated the impact of the layer normalization positions~\cite{wang-etal-2019-learning-deep,DBLP:conf/icml/XiongYHZZXZLWL20}.
There are currently two major layer normalization positions in Transformers: Pre-Layer Normalization (Pre-LN) and Post-Layer Normalization (Post-LN).
Pre-LN applies the layer normalization to an input for each sub-layer, and Post-LN places the layer normalization after each residual connection.
The original Transformer~\cite{NIPS2017_7181} employs Post-LN.
However, many recent studies have suggested using Pre-LN~\cite{wang-etal-2019-learning-deep,DBLP:journals/corr/abs-1809-10853,NEURIPS2020_1457c0d6} because the training of deep Transformers (e.g., those with ten or more layers) using Post-LN is often unstable, resulting in useless models.
Figure~\ref{fig:loss4encdec} shows loss curves for an actual example; the training of 18 layered Transformer encoder-decoders (18L-18L) on a widely used WMT English-to-German machine translation dataset.
These figures clearly show that the Post-LN Transformer encoder-decoders fail to train the model.
However, in contrast, \citet{liu-etal-2020-understanding} reported that Post-LN consistently achieved better performance than Pre-LN on a machine translation task when they used 6 layered (relatively shallow, 6L-6L) Transformers.

\begin{figure*}[!t]
  \centering 
  \includegraphics[width=16cm]{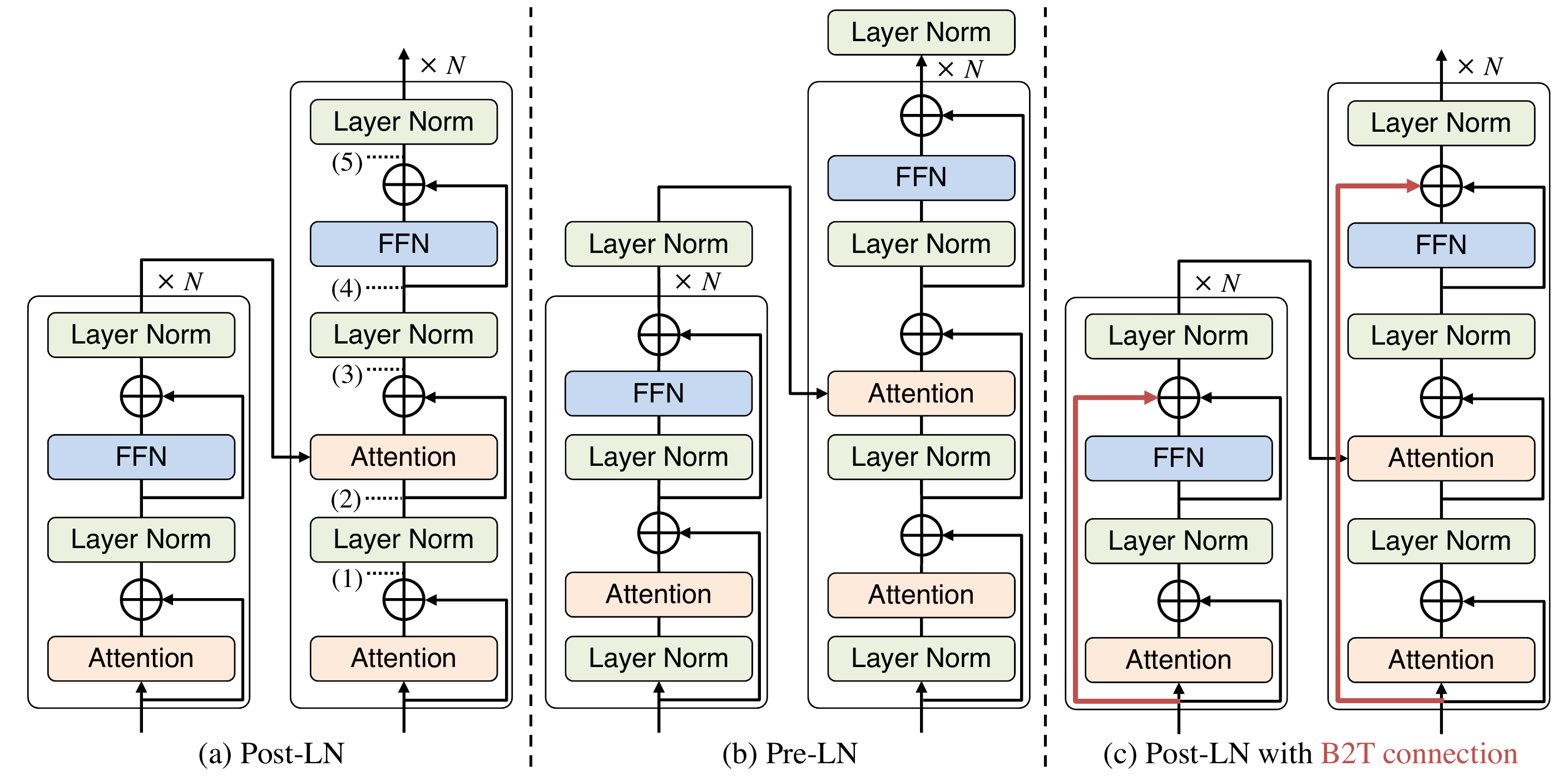}
   \caption{Transformer-based encoder-decoder architectures for (a) Post-LN, (b) Pre-LN, and (c) Post-LN with our proposed B2T connection.}
   \label{fig:overview}
\end{figure*}

This paper focuses specifically on such discrepancies between Pre-LN and Post-LN in configurations with various number of layers.
We investigate the sources of the instability of training in deep configurations and the superior performance in shallow configurations for Post-LN, compared with that for Pre-LN, to understand the essentials of the differences between Pre-LN and Post-LN.
We discover that the layer normalization in Post-LN is the main source of the vanishing gradient problem that leads to unstable training, whereas Pre-LN prevents it, as shown in Figure \ref{fig:loss4encdec}.
In particular, we clarify that the layer normalization is a significant factor of the vanishing gradient problem by comparing the input/output vector norms of gradient flows for each layer normalization during back-propagation.
These analyses bring us a novel idea that can satisfy higher stability by skipping over layer normalizations and provide better performance than Pre-LN regardless of their layer sizes.
Consequently, we propose a method that is based on Post-LN Transformers but has additional residual connections to enable stable training.

We conduct experiments on a wide range of text generation tasks, namely machine translation, summarization, language modeling, and ASR.
The experimental results lead to the following three new major findings:
\begin{enumerate}
  \item Post-LN Transformers achieve better performance than Pre-LN Transformers on text generation tasks (not only machine translation~\cite{liu-etal-2020-understanding} but also other tasks).
  Thus, Post-LN is superior to Pre-LN if the problem of its unstable training can be solved.
  \item Our modification enables Post-LN Transformers to stack many layers.
  \item Our method can maintain the performance advantage of Post-LN and mitigate its unstable training property, thus providing better performance than Pre-LN.
\end{enumerate}

\section{Post-LN and Pre-LN Transformers}
We briefly describe Post-LN and Pre-LN Transformers in this section.
The original Transformer~\cite{NIPS2017_7181} uses Post-LN, in which layer normalizations are located after each residual connection.
Let $x$ be an input of a sub-layer, and $\mathcal{F}(\cdot)$ be a sub-layer of a Transformer, such as a feed-forward network or multi-head attention.
Post-LN is defined as follows:
\begin{align}
\mathrm{PostLN}(x) = \mathrm{LN}(x + \mathcal{F}(x)), \label{eq:postln}
\end{align}
where $\mathrm{LN}(\cdot)$ is the layer normalization function.

In contrast, Pre-LN places the layer normalization before an input of each sub-layer:
\begin{align}
\mathrm{PreLN}(x) = x + \mathcal{F}(\mathrm{LN}(x)). \label{eq:preln}
\end{align}
Figure \ref{fig:overview} (a) and (b) illustrate Post-LN and Pre-LN Transformer architectures, respectively.

\section{Gradients of Transformer Layers}
\label{sec:gradient}

\begin{figure}[!t]
  \centering 
    \includegraphics[width=8cm]{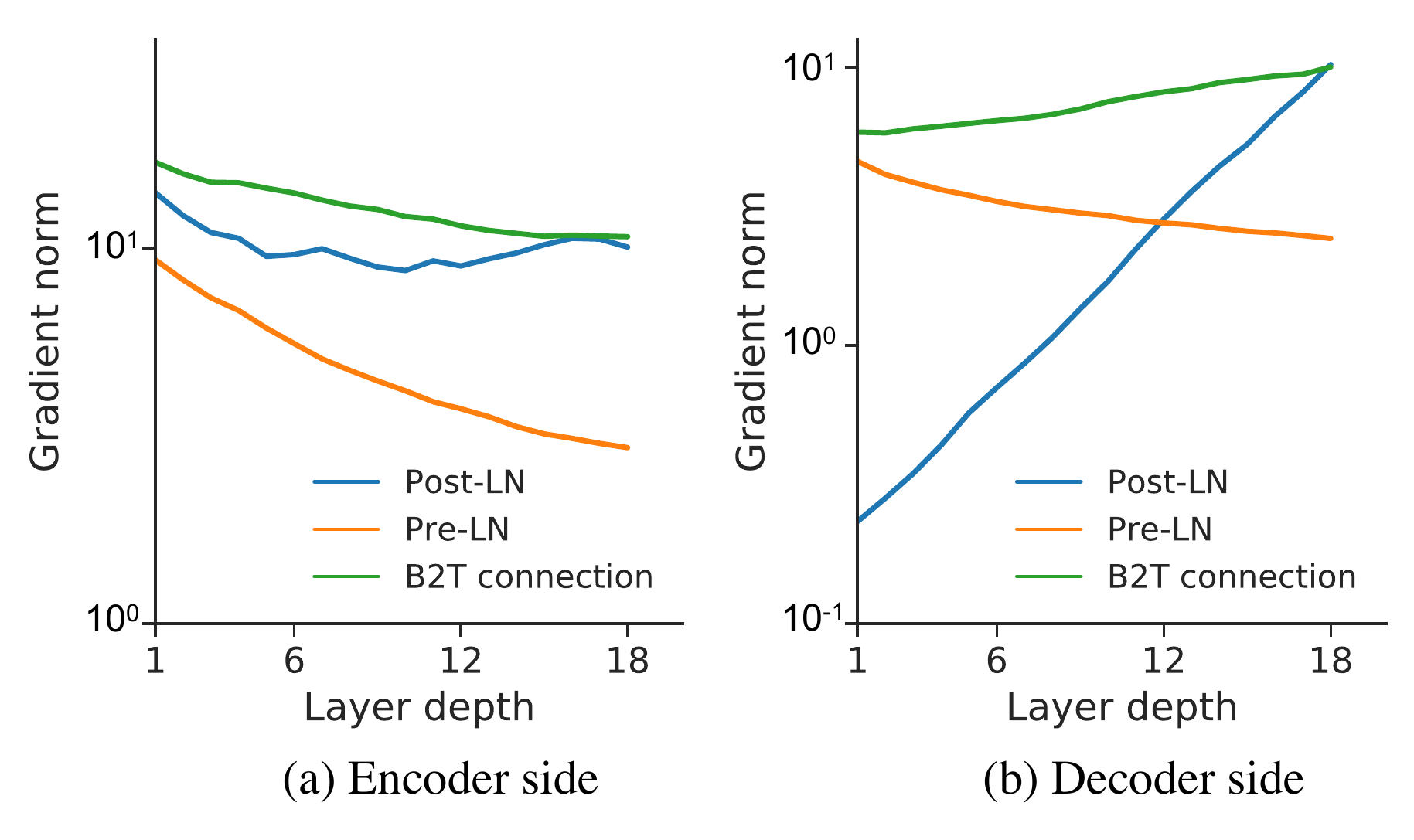}
   \caption{Gradient norms of 18 layered Transformer-based encoder-decoder architectures.}
   \label{fig:grad4encdec}
\end{figure}

\begin{figure}[!t]
  \centering 
  \includegraphics[width=8cm]{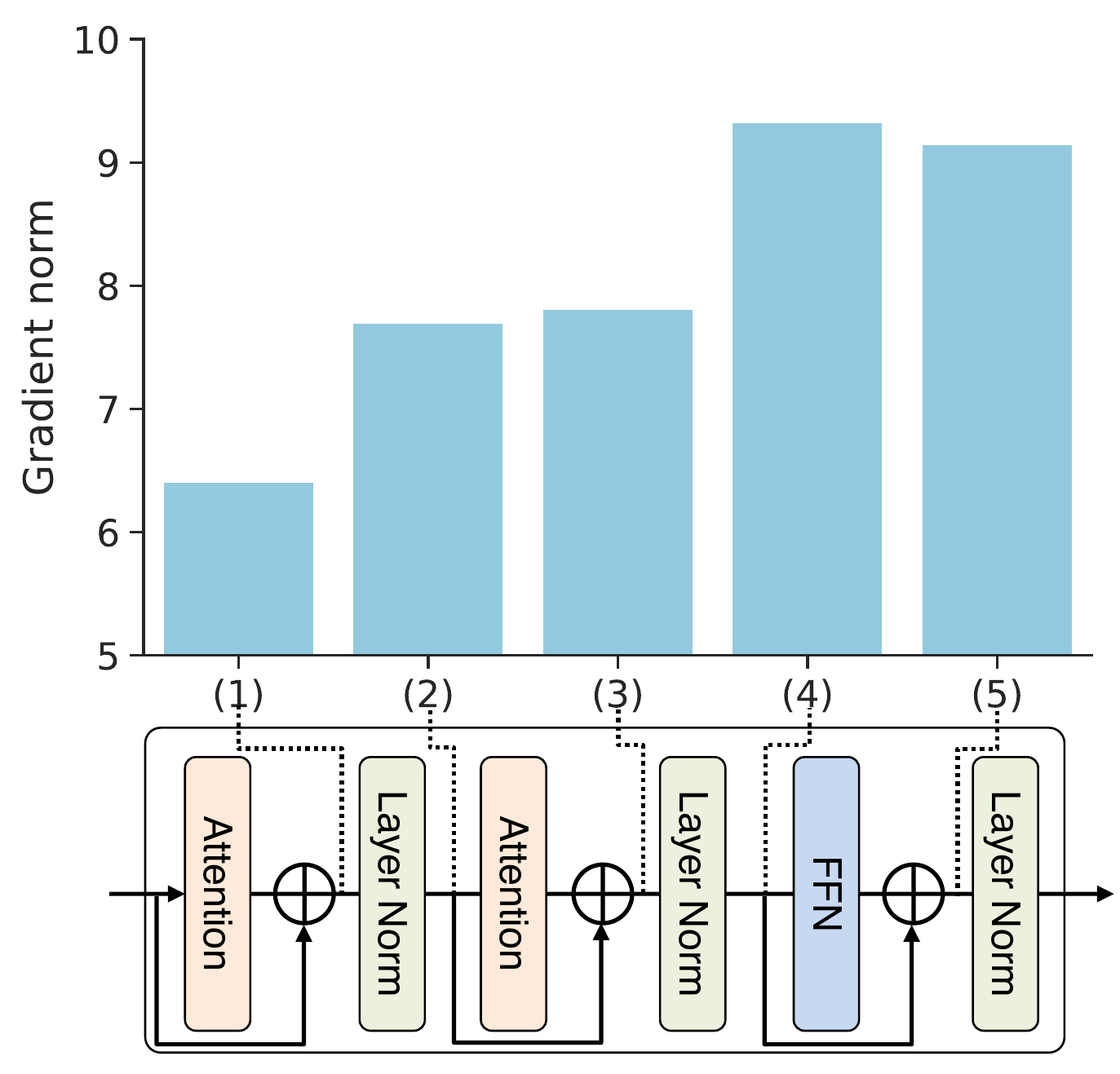}
   \caption{Gradient norms of each location in the 18th decoder for the 18 layered Post-LN Transformer encoder-decoder on WMT English-to-German translation training data.}
   \label{fig:grad4eachpart}
\end{figure}

As described in \citet{liu-etal-2020-understanding}, the vanishing gradient problem often occurs in Post-LN Transformers.
Figure \ref{fig:grad4encdec} shows the gradient norms of each layer for the (a) encoder-side and (b) decoder-side at the beginning of training, when 18L-18L Transformer encoder-decoders are trained on a widely used machine translation dataset (the WMT English-to-German dataset).
Focus on the decoder-side of Post-LN as illustrated in Figure \ref{fig:grad4encdec} (b).
This figure shows that shallower layers have smaller gradient norms.
In other words, the vanishing gradient occurs in the decoder-side of Post-LN because its gradient norms exponentially decay as they are back-propagated to shallower layers.
This result is consistent with the previous study~\cite{liu-etal-2020-understanding}.
We consider that this vanishing gradient causes the difficulty of stacking many layers with the Post-LN setting, as shown in Figure \ref{fig:loss4encdec}.

To investigate the vanishing gradient empirically in more detail, we measure the gradient norms of parts (1) - (5) of Figure \ref{fig:overview} (a).
Figure \ref{fig:grad4eachpart} shows the gradient norms of each part in the 18th layer\footnote{Appendix \ref{sec:appendix_grad} shows the gradient norms of each part in the 1st and 9th decoders as additional examples.}.
This figure shows that the gradient norms decrease drastically from (4) to (3) and (2) to (1).
These parts correspond to layer normalizations, as shown in Figure \ref{fig:grad4eachpart}.
This suggests that layer normalizations in Post-LN Transformers are probably the cause of the vanishing gradient problem.

To investigate the difference between the gradient flows of Post-LN and those of Pre-LN theoretically, we calculate the derivatives of Equations (\ref{eq:postln}) and (\ref{eq:preln}), as follows:
\begin{align}
\frac{\partial \mathrm{PostLN}(x)}{\partial x} &= \frac{\partial \mathrm{LN}(x + \mathcal{F}(x))}{\partial (x + \mathcal{F}(x))}\left(I + \frac{\partial \mathcal{F}(x)}{\partial x} \right), \label{eq:deriv_postln} \\
\frac{\partial \mathrm{PreLN}(x)}{\partial x} &= I + \frac{\partial \mathcal{F}(\mathrm{LN}(x))}{\partial \mathrm{LN}(x)}\frac{\partial \mathrm{LN}(x)}{\partial x}, \label{eq:deriv_preln}
\end{align}
where $I$ is the identity matrix.
As Equation (\ref{eq:deriv_postln}), the derivative of Post-LN is equal to the product of two derivatives: one is the layer normalization, and the other consists of the residual connection and sub-layer $\mathcal{F}$.
In contrast, in Pre-LN, the derivative of the residual connection is isolated from the term related to the derivative of the layer normalization.
The difference between these equations implies that the residual connection in Pre-LN prevents the vanishing gradient because it retains the gradients of upper layers even if the derivative of the layer normalization decreases gradients drastically.

\section{Transformations by Each Layer}
\label{sec:forward}

As described, it is difficult to stack many layers in Post-LN Transformers because the vanishing gradient problem occurs.
Although Pre-LN is more stable in training, Post-LN can achieve better performance if training succeeds (see Section \ref{sec:experiments}).
In this section, we explore the reason for this difference in performance.

Focus Pre-LN in Figure \ref{fig:grad4encdec}.
In contrast to Post-LN, in Pre-LN, a deeper (higher) layer has a smaller gradient norm.
Thus, the parameters of higher layers are not required to change dramatically from their initial values.
This implies that higher layers in Pre-LN are not sufficiently effective.

\begin{figure}[!t]
  \centering 
  \includegraphics[width=8cm]{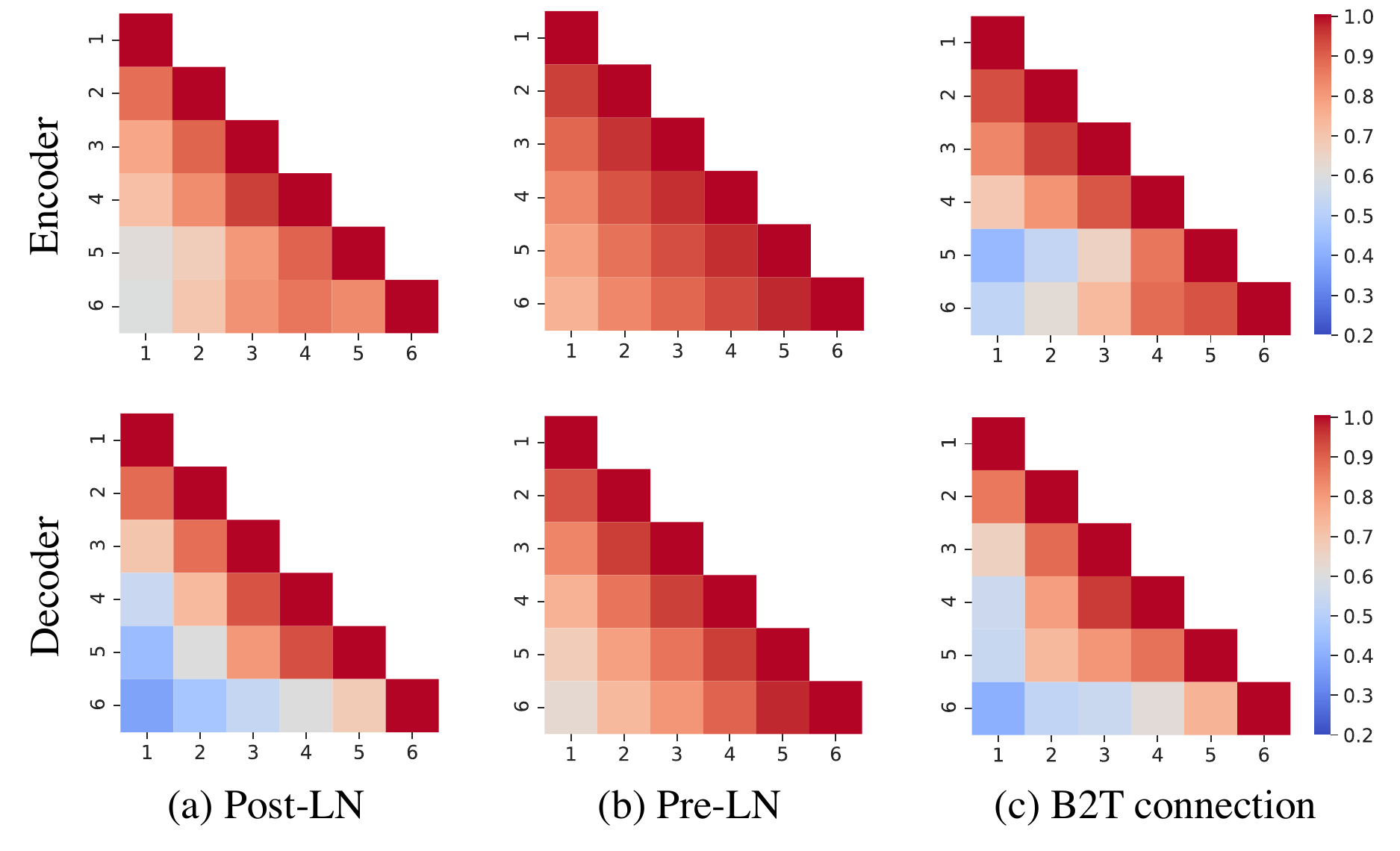}
   \caption{Cosine similarities between the outputs of each pair of layers.}
   \label{fig:heatmap}
\end{figure}

To investigate the effectiveness of higher layers, we focus on the transformations by each layer.
Figure \ref{fig:heatmap} shows the average cosine similarities between the outputs of each pair of layers for 6L-6L Transformer encoder-decoders trained on the WMT dataset when several sequences are input.
This figure indicates that the lower-left similarities of Pre-LN are higher than those of Post-LN.
This result means that the outputs of shallow layers are similar to the output of the final layer in Pre-LN, but not in Post-LN.
Consequently, higher layers in Pre-LN are less effective than those in Post-LN if training succeeds.

We consider that the residual connection in Pre-LN causes this phenomenon.
As Equation (\ref{eq:preln}) shows, in Pre-LN, an input $x$ skips over the sub-layer $\mathcal{F}(\cdot)$ by the residual connection.
Thus, the input $x$ is directly connected to the final layer output.
This property makes the training stable, as described in Section \ref{sec:gradient}, but causes high similarities between the outputs of the various layers.
Therefore, we consider that Pre-LN underperforms Post-LN because the residual connection in Pre-LN reduces the effectiveness of its higher layers.
In contrast, in Post-LN, larger gradient norms in higher layers (as shown in Figure \ref{fig:grad4encdec}) make higher layers more effective (as shown in Figure \ref{fig:heatmap}) but it is necessary to prevent the vanishing gradient problem in shallow layers when we stack many layers.

\section{Modification for Stable Training in Post-LN: Bottom-to-Top Connection}
This section introduces a modification that makes the training of Post-LN more stable while preserving its high performance.
This modification comprises an additional residual connection to mitigate the vanishing gradient in Post-LN by enabling many layers to be stacked.

As discussed in the previous sections, we need a term that retains gradients in the derivatives, as in Equation (\ref{eq:deriv_preln}), to prevent the vanishing gradient.
To satisfy this requirement, we propose a residual connection that skips over all layer normalizations except the final one in each layer.
Our introduced connection ties an input of a layer to the result of the feed-forward network (FFN), as illustrated by the red arrows in Figure \ref{fig:overview} (c).
We call this connection \textbf{Bottom-to-Top (B2T)} connection, which is formalized in the following equation:
\begin{align}
x_{inp} + x_{ffn} + \mathrm{FFN}(x_{ffn}), \label{eq:b2t}
\end{align}
where $x_{inp}$ is an input of a layer, $\mathrm{FFN}(\cdot)$ is an FFN, and $x_{ffn}$ is an input of the FFN.
In short, $x_{inp}$ skips the layer normalizations after the self-attention and encoder-decoder cross-attention.
Because the derivative of $x_{inp}$ is isolated from the terms related to the derivatives of the layer normalizations just behind the attention sub-layers, it retains gradients, as in Pre-LN.
For example, in an encoder-side, $x_{ffn}$ is as follows:
\begin{align}
  x_{ffn} = \mathrm{LN}(\mathrm{SelfAttn}(x_{inp}) + x_{inp}),
\end{align}
where $\mathrm{SelfAttn}(\cdot)$ is a self-attention network.
Thus, Equation (\ref{eq:b2t}) can be written as follows:
\begin{multline}
x_{inp} + \mathrm{LN}(\mathrm{SelfAttn}(x_{inp}) + x_{inp}) \\
+ \mathrm{FFN}(\mathrm{LN}(\mathrm{SelfAttn}(x_{inp}) + x_{inp})), \label{eq:b2t_for_encside}
\end{multline}
The derivative of this equation is the following equation:
\begin{multline}
I + \frac{\partial(\mathrm{LN}(\mathrm{SelfAttn}(x_{inp}) + x_{inp}))}{\partial x_{inp}} \\ + \frac{\partial(\mathrm{FFN}(\mathrm{LN}(\mathrm{SelfAttn}(x_{inp}) + x_{inp})))}{\partial x_{inp}}, 
\end{multline}
Because this derivative contains $I$, which is unrelated to the derivatives of internal layer normalizations, our B2T connection (i.e., $x_{inp}$) helps to propagate gradients.
For a decoder-side, we can prove this property in the same manner.

Figure \ref{fig:grad4encdec} (b) indicates that B2T connection mitigates the vanishing gradient of 18L-18L encoder-decoders.
Moreover, we locate B2T connection before the final layer normalization in each layer to avoid a direct connection to the final layer output based on the discussion in Section \ref{sec:forward}.
Thus, B2T connection preserves the property of Post-LN with respect to the transformations performed by each layer, as illustrated in Figure \ref{fig:heatmap} (c)\footnote{We also tried a connection that skips over all layer normalizations including the final one in each layer but it significantly impaired the performance. When we prepare such a connection, the connection ties an input to the output directly. Because this connection inhibits transformations performed by each layer as described in Section \ref{sec:forward}, it is reasonable that the performance is impaired. Therefore, we avoid skipping the final layer normalization in each layer to take the advantage of Post-LN.}.

\section{Experiments}
\label{sec:experiments}
Through experiments, we indicate following three findings.
\begin{itemize}
  \item Post-LN Transformers achieve better performance than Pre-LN Transformers if their training succeeds.
  \item B2T connection enables the training of deep Transformers with the Post-LN configuration.
  \item Our modification preserves the performance advantage of Post-LN Transformers, which therefore outperform Pre-LN Transformers.
\end{itemize}
We describe the essential experimental configurations in this section.
Appendix \ref{sec:detail_setting} presents more details, such as the hyper-parameters and computational budgets.

\begin{table*}[!t]
  \centering{}
  \footnotesize
  \begin{tabular}{ l | c c c c c c c | c} \hline
  Method & 2010 & 2011 & 2012 & 2013 & 2014 & 2015 & 2016 & Average \\ \hline \hline
  \multicolumn{9}{c}{Enc-Dec: 6L-6L} \\ \hline \hline
  Post-LN & 24.27 & \textbf{22.06} & \textbf{22.43} & 26.11 & 27.13 & 29.70 & 34.40 & \textbf{26.59} \\
  Pre-LN & 24.03 & 21.77 & 22.08 & 25.63 & 26.27 & 29.07 & 33.84 & 26.10 \\
  DLCL \cite{wang-etal-2019-learning-deep} & 23.94 & 22.00 & 22.24 & 26.11 & \textbf{27.37} & \textbf{29.71} & 34.26 & 26.52 \\
  Admin \cite{liu-etal-2020-understanding} & \textbf{24.32} & 21.79 & 22.17 & 26.26 & 27.14 & 29.61 & 34.12 & 26.49 \\
  T-Fixup \cite{pmlr-v119-huang20f} & 24.09 & 21.98 & 22.04 & 25.96 & 26.92 & 29.45 & 34.56 & 26.43 \\
  RealFormer \cite{he-etal-2021-realformer} & 24.18 & 22.02 & 22.17 & 26.02 & 26.98 & 29.36 & 34.15 & 26.41 \\
  DeepNet \cite{https://doi.org/10.48550/arxiv.2203.00555} & 24.08 & 21.76 & 22.09 & 25.90 & 26.85 & 29.62 & 34.39 & 26.38 \\
  B2T connection & 24.12 & 21.93 & 22.29 & \textbf{26.31} & 26.84 & 29.48 & \textbf{34.73} & 26.53 \\ \hline \hline
  \multicolumn{9}{c}{Enc-Dec: 18L-18L} \\ \hline \hline
  Post-LN & \multicolumn{7}{c|}{Training failed (See Figure \ref{fig:loss4encdec})} & N/A \\
  Pre-LN & 24.07 & 21.98 & 22.40 & 26.28 & 27.36 & 29.74 & 34.16 & 26.57 \\
  DLCL \cite{wang-etal-2019-learning-deep} & 24.20 & \textbf{22.51} & 22.83 & 26.59 & 27.97 & 30.24 & 33.98 & 26.90 \\
  Admin \cite{liu-etal-2020-understanding} & 24.56 & 22.17 & 22.62 & 26.48 & 27.99 & 30.35 & 33.88 & 26.86 \\
  T-Fixup \cite{pmlr-v119-huang20f} & 24.45 & 22.29 & 22.76 & 26.57 & 27.71 & 30.13 & 34.69 & 26.94 \\
  RealFormer \cite{he-etal-2021-realformer} & 24.32 & 22.42 & 22.68 & 26.59 & \textbf{28.58} & 30.36 & 33.71 & 26.95 \\
  DeepNet \cite{https://doi.org/10.48550/arxiv.2203.00555} & \textbf{24.70} & 22.40 & \textbf{22.92} & \textbf{26.85} & 28.21 & 30.60 & 34.25 & 27.13 \\
  B2T connection & 24.62 & \textbf{22.51} & 22.86 & 26.74 & 28.48 & \textbf{30.99} & \textbf{34.93} & \textbf{27.30} \\ \hline
  \end{tabular}
  \caption{BLEU scores of each method on WMT newstest2010-2016 and their averages.\label{tab:exp_mt}}
\end{table*}

\subsection{Machine Translation}
\label{sec:exp_mt}

\subsubsection{Dataset}
\label{sec:mt_dataset}
The machine translation task has been widely used to investigate the performance of Transformer-based methods since the original Transformer~\cite{NIPS2017_7181,ott-etal-2018-scaling,wang-etal-2019-learning-deep,DBLP:conf/icml/XiongYHZZXZLWL20,liu-etal-2020-understanding}.
We adopted the widely used WMT English-to-German training dataset~\cite{NIPS2017_7181,ott-etal-2018-scaling}, which contains 4.5M sentence pairs.
We applied the byte-pair-encoding (BPE) algorithm~\cite{sennrich-etal-2016-neural} to construct a vocabulary set in the same manner as previous studies.
We set the number of BPE merge operations to 32K and shared the vocabulary between the source and target languages.
We used newstest2010-2016 to investigate the performance, following \citet{takase-kiyono-2021-rethinking}.

\subsubsection{Methods}
\label{sec:mt_method}
We compare \textbf{Post-LN}, \textbf{Pre-LN}, and Post-LN with our B2T connection (\textbf{B2T connection}) Transformers.
We used \texttt{fairseq}\footnote{\href{https://github.com/pytorch/fairseq}{https://github.com/pytorch/fairseq}} \cite{ott-etal-2019-fairseq} as an implementation of Transformers.
We stacked 6 and 18 layers for the encoders and decoders (6L-6L and 18L-18L) as the widely used configuration and deep configuration, respectively.
We used the Transformer (base) setting for dimension sizes of internal layers.
In addition to the above methods, we evaluate the following five methods, which are recent approaches that enable the training of deep Transformers.
We used the same hyper-parameters for all methods except T-Fixup.
For T-Fixup, we used the hyper-parameters reported in \citet{pmlr-v119-huang20f} to prevent divergence.

\noindent
\textbf{DLCL}
To make Transformers deep, \citet{wang-etal-2019-learning-deep} proposed dynamic linear combination of layers (DLCL), which uses the weighted sum of the lower layers as an input of a layer.
In contrast to our B2T connection, which is an additional connection within each layer, DLCL uses a connection among layers.
We apply DLCL to Post-LN Transformers.
We used the official implementation\footnote{\href{https://github.com/wangqiangneu/dlcl}{https://github.com/wangqiangneu/dlcl}}.

\noindent
\textbf{Admin}
\citet{liu-etal-2020-understanding} proposed adaptive model initialization (Admin), which uses additional parameters to stabilize the training of Post-LN Transformers.
This method requires the variances of internal layers to initialize the additional parameters.
Thus, this method first processes several forward steps for the initialization, and then conducts the actual training.
In a nutshell, this method incurs additional computational costs.
We used the official implementation\footnote{\href{https://github.com/LiyuanLucasLiu/Transformer-Clinic}{https://github.com/LiyuanLucasLiu/Transformer-Clinic}}.

\noindent
\textbf{T-Fixup}
\citet{pmlr-v119-huang20f} proposed an initialization scheme for Transformers, T-Fixup, to perform stable training without the learning rate warm-up and layer normalizations.
Because this method can remove the cause of the vanishing gradient, we can stack many layers.
We used the official implementation\footnote{\href{https://github.com/layer6ai-labs/T-Fixup}{https://github.com/layer6ai-labs/T-Fixup}}.

\noindent
\textbf{RealFormer}
To improve the performance of Transformers, \citet{he-etal-2021-realformer} proposed RealFormer, which introduces additional connections into attention sub-layers.
Although their motivation is not addressing the vanishing gradient problem, their method is similar to ours with respect to the use of additional connections.

\noindent
\textbf{DeepNet}
\citet{https://doi.org/10.48550/arxiv.2203.00555} proposed DeepNorm, which uses a weight that corresponds to the number of layers in a residual connection before layer normalizations to stabilize Post-LN based Transformers.
They also provided the combination of the initialization scheme and DeepNorm as DeepNet.

\subsubsection{Results}
We measured case-sensitive detokenized BLEU scores with SacreBLEU~\cite{post-2018-call}\footnote{The BLEU scores calculated by SacreBLEU are often lower than those calculated by the procedure of \citet{NIPS2017_7181} as reported in \citet{ott-etal-2018-scaling}. In fact, Pre-LN and B2T connection in the 18L-18L configuration achieved scores of 28.94 and 29.91, respectively, on newstest2014 when we used the same procedure of \citet{NIPS2017_7181}. However, we used SacreBLEU to ensure the compatibility of results, as described in \citet{post-2018-call}.}.
Table \ref{tab:exp_mt} shows BLEU scores\footnote{The signature of SacreBLEU is BLEU+nrefs:1+ case:mixed+eff:no+tok:13a+smooth:exp+version:2.0.0.} of each method on newstest2010-2016 and the averaged scores of them.
Since the BLEU score is precision-based n-gram overlapping between the model output and correct examples, a higher score represents better performance.

The upper part of Table \ref{tab:exp_mt} shows results in the 6L-6L configuration.
This part indicates that Post-LN achieved better scores than Pre-LN on all test sets.
In addition, B2T connection outperformed Pre-LN on all test sets.
Thus, these methods are superior to Pre-LN when the total number of layers is small.

\begin{figure}[!t]
  \centering 
  \includegraphics[width=7.5cm]{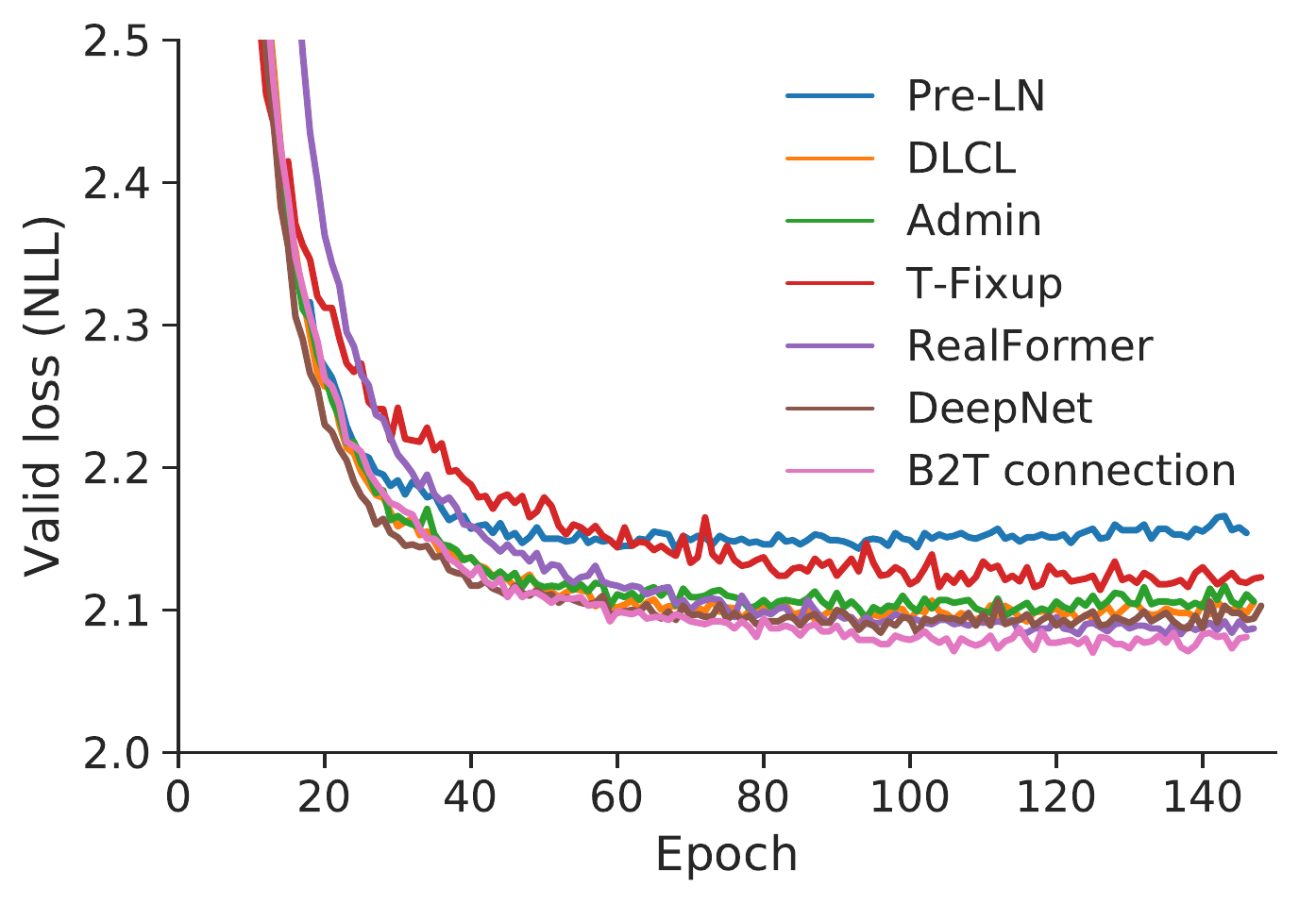}
   \caption{Negative Log-Likelihood (NLL) on validation data (newstest2013) when we stack 18 layers.}
   \label{fig:training_curve}
\end{figure}

\begin{table*}[!t]
  \centering{}
  \footnotesize
  \begin{tabular}{ l | c c c c c c c | c} \hline
  Method & 2010 & 2011 & 2012 & 2013 & 2014 & 2015 & 2016 & Average \\ \hline \hline
  \multicolumn{9}{c}{Enc-Dec: 100L-100L} \\ \hline \hline
  Post-LN & \multicolumn{7}{c|}{Training failed} & N/A \\
  Pre-LN & 24.81 & 22.67 & 23.15 & 26.98 & 28.42 & 30.50 & 34.53 & 27.29 \\
  B2T connection & \textbf{25.26} & \textbf{23.27} & \textbf{23.72} & \textbf{27.50} & \textbf{29.33} & \textbf{31.57} & \textbf{35.37} & \textbf{28.00} \\ \hline
  \end{tabular}
  \caption{BLEU scores on WMT newstest2010-2016 and their averages in the 100L-100L configuration.\label{tab:exp_mt_deep}}
\end{table*}

The lower part of Table \ref{tab:exp_mt} shows results in the 18L-18L configuration.
This part shows that the training of Post-LN failed, and thus we cannot successfully stack 18L-18L in the vanilla Post-LN.
With the B2T connection, its training succeeded and it outperformed Pre-LN in the 18L-18L configuration.
Figure \ref{fig:training_curve} shows the negative log-likelihood (NLL) values of all methods when we regard newstest2013 as validation data.
This figure indicates that the NLLs of Pre-LN are worse than those of the other methods.
These results demonstrate that our modification enabled the stacking of many layers without harm to its performance such as Pre-LN.

In the comparison with the recent methods, B2T connection outperformed them with respect to the averaged BLEU score.
This result implies that our modification is superior to the recent methods.
To make our findings more reliable, we also conduct a comparison with the recent methods on the summarization task.

Table \ref{tab:exp_mt_deep} shows results in a much deeper configuration: 100L-100L.
This table also indicates that B2T connection stabilized the training and outperformed Pre-LN.
Appendix \ref{sec:stack100layer} describes the details of this 100L-100L configuration and shows a comparison with the latest method, DeepNet~\cite{https://doi.org/10.48550/arxiv.2203.00555}.

\subsection{Abstractive Summarization}
\label{sec:exp_summarization}
\subsubsection{Dataset}
The abstractive summarization task is one of the most famous sequence-to-sequence problems in NLP.
In this study, we conduct the experiment on the headline generation task, which is the task of generating a headline from a given sentence~\cite{rush-chopra-weston:2015:EMNLP}.
We used headline-sentence pairs extracted from Annotated English Gigaword~\cite{napoles:2012:AG} by \citet{rush-chopra-weston:2015:EMNLP}.
This dataset contains 3.8M headline-sentence pairs as the training set and 1951 pairs as the test set.
In addition, we used 13M additional headline-sentence pairs extracted from REALNEWS~\cite{NEURIPS2019_3e9f0fc9} and NewsCrawl~\cite{barrault-etal-2019-findings} for training deep Transformers, following \citet{takase-kiyono-2021-rethinking}.
We applied BPE~\cite{sennrich-etal-2016-neural} to construct a vocabulary set.
As in the machine translation experiments, we set the number of BPE merge operations to 32K and shared the vocabulary between the encoder and decoder sides.

\subsubsection{Methods}
We compare \textbf{Post-LN}, \textbf{Pre-LN}, and \textbf{B2T connection} Transformers in the same manner as in Section \ref{sec:exp_mt}.
In addition, we compare \textbf{DLCL}, \textbf{Admin}, \textbf{T-Fixup}, \textbf{RealFormer}, and \textbf{DeepNet} because it would be premature to conclude that our modification is more effective than those methods from the results of experiments on the machine translation task alone.
We set the numbers of layers of encoders and decoders to 6L-6L and 18L-18L as the base and deep configurations, respectively.

\subsubsection{Results}
\begin{table}
  \centering{}
  \footnotesize
  \begin{tabular}{ l | c c c } \hline
  Method & R-1 & R-2 & R-L \\ \hline \hline
  \multicolumn{4}{c}{Enc-Dec: 6L-6L} \\ \hline \hline
  Post-LN & \textbf{38.57} & \textbf{19.37} & \textbf{35.79} \\
  Pre-LN & 38.27 & 19.29 & 35.39 \\
  DLCL \cite{wang-etal-2019-learning-deep} & 38.13 & 18.49 & 35.00 \\
  Admin \cite{liu-etal-2020-understanding} & 37.96 & 18.93 & 35.05 \\
  T-Fixup \cite{pmlr-v119-huang20f} & 38.11 & 19.13 & 35.32 \\
  RealFormer \cite{he-etal-2021-realformer} & 38.30 & 19.32 & 35.46 \\
  DeepNet \cite{https://doi.org/10.48550/arxiv.2203.00555} & 38.27 & 18.89 & 35.34 \\
  B2T connection & 38.43 & \textbf{19.37} & 35.72 \\ \hline \hline
  \multicolumn{4}{c}{Enc-Dec: 18L-18L} \\ \hline \hline
  Post-LN & \multicolumn{3}{c}{Training failed} \\
  Pre-LN & 38.97 & 19.94 & 35.99 \\
  DLCL \cite{wang-etal-2019-learning-deep} & 38.25 & 19.44 & 35.57 \\
  Admin \cite{liu-etal-2020-understanding} & 39.10 & 20.08 & 36.30 \\
  T-Fixup \cite{pmlr-v119-huang20f} & 39.15 & 19.97 & 36.34 \\
  RealFormer \cite{he-etal-2021-realformer} & 39.22 & 20.12 & 36.49 \\
  DeepNet \cite{https://doi.org/10.48550/arxiv.2203.00555} & 39.27 & 19.97 & 36.41 \\
  B2T connection & \textbf{39.61} & \textbf{20.28} & \textbf{36.66} \\ \hline
  \end{tabular}
  \caption{F1 based ROUGE-1, 2, and L scores (columns headed R-1, R-2, and R-L, respectively) on headline generation~\cite{rush-chopra-weston:2015:EMNLP}.\label{tab:exp_summarization}}
\end{table}

Table \ref{tab:exp_summarization} shows the ROUGE-1, 2, and L scores achieved by each method on the test set.
Since these scores are computed by n-gram overlapping between the generated and correct headlines, a higher score represents better performance.

In the 6L-6L configuration, Post-LN achieved better performance than Pre-LN.
Thus, Post-LN outperformed Pre-LN on the headline generation task if training succeeded.
Moreover, B2T connection achieved scores comparable to those of Post-LN.

In the 18L-18L configuration, the training of Post-LN failed.
In contrast, the training of B2T connection succeeded, and this method outperformed Pre-LN.
Thus, our modification is more suitable than Pre-LN for training deep Transformers to perform the headline generation task.

B2T connection outperformed the recent methods in the 6L-6L configuration and achieved the best ROUGE scores in the 18L-18L configuration.
According to the results on both the machine translation and headline generation tasks, B2T connection achieved performance that was better than, or comparable to, that of previous methods.
It is worth emphasizing that, in addition to the performance, our modification does not incur additional computational costs, such as those incurred by DLCL and Admin.

\subsection{Language Model}
In addition to encoder-decoders, we investigate the effect of our B2T connection when used in the decoder side only, i.e., a neural language model.
Because recent pre-trained models, such as the GPT series, are language models trained on a large amount of training data, experimental results in this section give an insight for pre-trained models.

\subsubsection{Dataset}
We used WikiText-103~\cite{DBLP:journals/corr/MerityXBS16}, which consists of a large number of tokens.
The training, validation, and test sets contain 103M, 0.2M, and 0.2M tokens, respectively.
The vocabulary set contains 0.3M words.

\subsubsection{Methods}
We used a Transformer with adaptive input representations~\cite{DBLP:journals/corr/abs-1809-10853}, which is implemented in \texttt{fairseq}, as the base architecture in this experiment.
For the base configuration, we stacked 6 layers, in the same manner as in the machine translation and summarization experiments.
For the deep configuration, we used 16 layers, following \citet{DBLP:journals/corr/abs-1809-10853}.
For the dimensions of internal layers, we used the same values as those used by \citet{DBLP:journals/corr/abs-1809-10853}.
We compare Post-LN, Pre-LN, and B2T connection.

\subsubsection{Results}
\begin{table}
  \centering{}
  \footnotesize
  \begin{tabular}{ l | c c c } \hline
  Method & Valid & Test \\ \hline \hline
  \multicolumn{3}{c}{Dec: 6L} \\ \hline \hline
  Post-LN & \textbf{20.24} & \textbf{21.22} \\
  Pre-LN & 20.98 & 21.93 \\
  B2T connection & 20.50 & 21.47 \\ \hline \hline
  \multicolumn{3}{c}{Dec: 16L} \\ \hline \hline
  Post-LN & \multicolumn{2}{c}{Training failed} \\
  Pre-LN & 18.53 & 19.24 \\
  B2T connection & \textbf{18.38} & \textbf{19.20} \\ \hline
  \end{tabular}
  \caption{Perplexities on WikiText-103~\cite{DBLP:journals/corr/MerityXBS16}.\label{tab:exp_lm}}
\end{table}

Table \ref{tab:exp_lm} shows perplexities of each method on the validation and test sets of WikiText-103.
Since the perplexity is computed based on the negative log-likelihood, a smaller value corresponds to better performance.
The upper part of this table indicates that, with 6 layers, Post-LN and our B2T connection outperformed Pre-LN.
When we stacked 16 layers, the training of Post-LN failed, but B2T connection achieved better performance than Pre-LN.
These results are consistent with results on the machine translation and summarization tasks.
Thus, our modification enables the training of deep Transformers for language modeling, and it is more effective than Transformers with Pre-LN.

\subsection{Automatic Speech Recognition}
\label{sec:exp_asr}
In addition to experiments on natural language processing tasks, we conduct an experiment on another modality, ASR.

\subsubsection{Dataset}
We used LibriSpeech~\cite{7178964}, which is the standard English ASR benchmark dataset.
The dataset contains 1,000 hours of English speech extracted from audiobooks.
We used the standard splits of LibriSpeech: we used all available training data for training and two configurations (`clean' and `other') of development sets and test sets for evaluation.
We applied the same pre-processing as that used by \citet{wang-etal-2020-fairseq}.
We constructed a vocabulary set for the decoder-side with SentencePiece~\cite{kudo-richardson-2018-sentencepiece} by setting the vocabulary size to 10,000.
To obtain speech features, we used torchaudio\footnote{\href{https://github.com/pytorch/audio}{https://github.com/pytorch/audio}}.

\subsubsection{Methods}
\begin{table}[!t]
  \centering{}
  \footnotesize
  \begin{tabular}{ l | c c c c} \hline
  & \multicolumn{2}{c}{Dev} & \multicolumn{2}{c}{Test} \\
  Method & Clean & Other & Clean & Other \\ \hline \hline
  \multicolumn{5}{c}{Enc-Dec: 6L-6L} \\ \hline \hline
  Post-LN & 3.78 & \textbf{8.76} & 4.19 & \textbf{8.74} \\
  Pre-LN & 3.89 & 9.69 & 4.22 & 9.65 \\
  B2T connection & \textbf{3.69} & 8.97 & \textbf{3.86} & 8.94 \\ \hline \hline
  \multicolumn{5}{c}{Enc-Dec: 12L-6L} \\ \hline \hline
  Post-LN & \multicolumn{4}{c}{Training failed} \\
  Pre-LN & \textbf{3.21} & 7.91 & 3.49 & 8.22 \\
  B2T connection & 3.26 & \textbf{7.74} & \textbf{3.48} & \textbf{7.68} \\ \hline
  \end{tabular}
  \caption{Word error rates of each method on LibriSpeech.\label{tab:exp_asr}}
\end{table}
We used the Transformer-based speech-to-text model described in \citet{wang-etal-2020-fairseq} as the base architecture in this experiment.
This model contains a convolutional layer to construct an embedding for the encoder-side but the other parts are identical to the Transformers used on the machine translation and summarization tasks.
We used the same dimensions as those of T-Md, described in \citet{wang-etal-2020-fairseq}.
We set the numbers of layers to 6L-6L and 12L-6L as the base and deep configurations, respectively, because \citet{wang-etal-2020-fairseq} stacked many layers on the encoder-side only.
We compare Post-LN, Pre-LN, and B2T connection.

\subsubsection{Results}
Table \ref{tab:exp_asr} shows the word error rates (WERs) of each method on each set.
A smaller value of WER corresponds to better performance.
The upper part of this table indicates that Post-LN and B2T connection outperformed Pre-LN on all sets in the 6L-6L configuration.
The lower part of the table shows that B2T connection succeeded in training and achieved performance that was better than (or comparable to) that of Pre-LN in the 12L-6L configuration\footnote{\citet{wang-etal-2020-fairseq} reported that the improvement was small even if they increased the number of parameters. Thus, we emphasize that B2T connection achieved better WERs on dev-other and test-other even though the number of parameters of B2T connection is (almost) equal to that of Pre-LN.}.
These results are consistent with those of the other experiments in this study.

\section{Related Work}
\label{sec:related_work}
Layer normalization~\cite{ba2016layer} is a useful technique for training neural networks but its mechanism has been unclear~\cite{NEURIPS2019_2f4fe03d}.
The Transformer, which is the standard architecture for various tasks, also contains layer normalizations.
The original Transformer architecture adopted the Post-LN configuration~\cite{NIPS2017_7181}.
However, recent Transformer implementations have adopted Pre-LN configurations~\cite{klein-etal-2017-opennmt,vaswani-etal-2018-tensor2tensor,ott-etal-2019-fairseq,DBLP:journals/corr/abs-1809-10853}.

To construct deep Transformers that achieve better performance, recent studies have focused on the behavior of layer normalizations.
\citet{wang-etal-2019-learning-deep} indicated the difficulty of training deep Transformers with Post-LN due to the vanishing gradient problem, and demonstrated that Pre-LN enables the stacking of many layers through machine translation experiments.
In addition, they proposed a method to connect all layers to increase the effectiveness of deep Transformers.
\citet{bapna-etal-2018-training} and \citet{dou-etal-2018-exploiting} also proposed such connection methods to stack many layers.
\citet{he-etal-2021-realformer} introduced additional connections into attention sub-layers to improve the performance.
\citet{DBLP:conf/icml/XiongYHZZXZLWL20} explored the relation between the warm-up strategy and layer normalizations in Transformers.
Through theoretical and empirical analyses, they indicated that Post-LN requires the warm-up strategy to stabilize the training.

\citet{liu-etal-2020-understanding} analyzed the training dynamics of Post-LN and Pre-LN Transformers.
They then proposed Admin, which consists of additional weight parameters to control the variances of outputs from each sub-layer.
In contrast, we indicated that we can stabilize the training of Post-LN Transformers by adding only a residual connection that skips over layer normalizations that cause the vanishing gradient.

Some studies have proposed initialization methods to make the training of deep neural networks stable~\cite{zhang-etal-2019-improving,conf/iclr/ZhangDM19,pmlr-v119-huang20f}.
\citet{zhang-etal-2019-improving} proposed the depth-scaled initialization to prevent the vanishing gradient problem in Transformers.
\citet{conf/iclr/ZhangDM19} proposed the fixed-update initialization to remove normalizations in neural networks.
Inspired by these studies, \citet{pmlr-v119-huang20f} proposed T-Fixup, which enables both warm-up and layer normalizations to be removed from Transformers.
In addition to the initialization scheme, \citet{https://doi.org/10.48550/arxiv.2203.00555} introduced weights into residual connections before layer normalizations, following \citet{liu-etal-2020-understanding}.

\section{Conclusion}
In this study, we addressed the stability of training Post-LN Transformers.
Through theoretical and empirical analyses, we indicated that layer normalizations cause the unstable training when many layers are stacked.
In addition, we investigated the reason for the different performance of Pre-LN and Post-LN by transformations of each layer.
We introduced B2T connection to prevent the vanishing gradient while preserving the advantage of Post-LN.
We conducted experiments on various tasks.
The experimental results led to the following three findings; 1, Post-LN achieved better performance than Pre-LN if its training succeeded.
2, Our modification enabled the training of deep Transformers (e.g., those with ten or more layers).
3, Our modification preserved the benefit of Post-LN, and therefore outperformed Pre-LN.

\section*{Limitations}
\label{sec:limitations}
In this paper, we indicated that the vanishing gradient problem, caused by layer normalizations, makes the training of deep Post-LN Transformers unstable.
We proposed the B2T connection to mitigate this vanishing gradient problem.
However, the proposed B2T connection does not perfectly prevent the vanishing gradient, as shown in Figure \ref{fig:grad4encdec}.
Therefore, the vanishing gradient might harm the training in extremely deep Transformers even if our B2T connection is used.

In addition, this study depends on empirical observations.
In particular, we provided little theoretical justification of the reason for Post-LN outperforming Pre-LN when training succeeds.
However, as discussed in Appendix \ref{sec:stack100layer}, the method with a theoretical justification often collapses in some situations.
Because the behavior of deep Transformers in various situations is not fully understood, we believe that it is important to provide empirical findings for our research field to progress.

Although Appendix \ref{sec:stack100layer} includes a comparison between our B2T connection and the latest method, DeepNet~\cite{https://doi.org/10.48550/arxiv.2203.00555}, we could not investigate the behavior of all methods in the 100L-100L configuration because of our limited computational budgets.
However, we are confident that we conducted sufficient experiments to verify our contributions.

\section*{Ethics Statement}
The proposed method helps to construct deep Transformers.
As discussed in \citet{strubell-etal-2019-energy} and \citet{DBLP:journals/corr/abs-1907-10597}, such deep neural networks consume substantial amounts of energy.
In fact, as discussed in Appendix \ref{sec:comp_res}, we spent a large amount of computational resources on our experiments.
Therefore, we also need to explore methods of improving energy efficiency while maintaining the good performance achieved by stacking many layers.

With respect to ethical considerations, the datasets used in our experiments are publicly available.
LibriSpeech~\cite{7178964} is derived from audiobooks.
The other datasets are mainly constructed from newswire texts and Wikipedia.
Thus, in our understanding, our used datasets do not contain any personally identifiable information or offensive contents.

\section*{Acknowledgements}
We thank the anonymous reviewers for their useful suggestions.
A part of this work was supported by JSPS KAKENHI Grant Number JP21K17800 and JST ACT-X Grant Number JPMJAX200I.
The work of Jun Suzuki was partly supported by JST Moonshot R\&D Grant Number JPMJMS2011 (fundamental research).
We thank \href{https://jp.edanz.com/ac}{Edanz} for editing a draft of this manuscript.

\bibliographystyle{acl_natbib}

\begin{thebibliography}{37}
\expandafter\ifx\csname natexlab\endcsname\relax\def\natexlab#1{#1}\fi

\bibitem[{Ba et~al.(2016)Ba, Kiros, and Hinton}]{ba2016layer}
Jimmy~Lei Ba, Jamie~Ryan Kiros, and Geoffrey~E. Hinton. 2016.
\newblock \href {http://arxiv.org/abs/1607.06450} {Layer normalization}.

\bibitem[{Baevski and Auli(2019)}]{DBLP:journals/corr/abs-1809-10853}
Alexei Baevski and Michael Auli. 2019.
\newblock Adaptive input representations for neural language modeling.
\newblock In \emph{Proceedings of the 7th International Conference on Learning
  Representations (ICLR)}.

\bibitem[{Bapna et~al.(2018)Bapna, Chen, Firat, Cao, and
  Wu}]{bapna-etal-2018-training}
Ankur Bapna, Mia Chen, Orhan Firat, Yuan Cao, and Yonghui Wu. 2018.
\newblock Training deeper neural machine translation models with transparent
  attention.
\newblock In \emph{Proceedings of the 2018 Conference on Empirical Methods in
  Natural Language Processing (EMNLP)}, pages 3028--3033.

\bibitem[{Barrault et~al.(2019)Barrault, Bojar, Costa-juss{\`a}, Federmann,
  Fishel, Graham, Haddow, Huck, Koehn, Malmasi, Monz, M{\"u}ller, Pal, Post,
  and Zampieri}]{barrault-etal-2019-findings}
Lo{\"\i}c Barrault, Ond{\v{r}}ej Bojar, Marta~R. Costa-juss{\`a}, Christian
  Federmann, Mark Fishel, Yvette Graham, Barry Haddow, Matthias Huck, Philipp
  Koehn, Shervin Malmasi, Christof Monz, Mathias M{\"u}ller, Santanu Pal, Matt
  Post, and Marcos Zampieri. 2019.
\newblock Findings of the 2019 conference on machine translation ({WMT}19).
\newblock In \emph{Proceedings of the Fourth Conference on Machine Translation
  (WMT)}, pages 1--61.

\bibitem[{Brown et~al.(2020)Brown, Mann, Ryder, Subbiah, Kaplan, Dhariwal,
  Neelakantan, Shyam, Sastry, Askell, Agarwal, Herbert-Voss, Krueger, Henighan,
  Child, Ramesh, Ziegler, Wu, Winter, Hesse, Chen, Sigler, Litwin, Gray, Chess,
  Clark, Berner, McCandlish, Radford, Sutskever, and
  Amodei}]{NEURIPS2020_1457c0d6}
Tom Brown, Benjamin Mann, Nick Ryder, Melanie Subbiah, Jared~D Kaplan, Prafulla
  Dhariwal, Arvind Neelakantan, Pranav Shyam, Girish Sastry, Amanda Askell,
  Sandhini Agarwal, Ariel Herbert-Voss, Gretchen Krueger, Tom Henighan, Rewon
  Child, Aditya Ramesh, Daniel Ziegler, Jeffrey Wu, Clemens Winter, Chris
  Hesse, Mark Chen, Eric Sigler, Mateusz Litwin, Scott Gray, Benjamin Chess,
  Jack Clark, Christopher Berner, Sam McCandlish, Alec Radford, Ilya Sutskever,
  and Dario Amodei. 2020.
\newblock Language models are few-shot learners.
\newblock In \emph{Advances in Neural Information Processing Systems 33
  (NeurIPS)}, pages 1877--1901.

\bibitem[{Dou et~al.(2018)Dou, Tu, Wang, Shi, and
  Zhang}]{dou-etal-2018-exploiting}
Zi-Yi Dou, Zhaopeng Tu, Xing Wang, Shuming Shi, and Tong Zhang. 2018.
\newblock Exploiting deep representations for neural machine translation.
\newblock In \emph{Proceedings of the 2018 Conference on Empirical Methods in
  Natural Language Processing (EMNLP)}, pages 4253--4262.

\bibitem[{He et~al.(2016{\natexlab{a}})He, Zhang, Ren, and Sun}]{7780459}
Kaiming He, Xiangyu Zhang, Shaoqing Ren, and Jian Sun. 2016{\natexlab{a}}.
\newblock Deep residual learning for image recognition.
\newblock In \emph{2016 IEEE Conference on Computer Vision and Pattern
  Recognition (CVPR)}, pages 770--778.

\bibitem[{He et~al.(2016{\natexlab{b}})He, Zhang, Ren, and
  Sun}]{10.1007/978-3-319-46493-0_38}
Kaiming He, Xiangyu Zhang, Shaoqing Ren, and Jian Sun. 2016{\natexlab{b}}.
\newblock Identity mappings in deep residual networks.
\newblock In \emph{14th European Conference on Computer Vision}, pages
  630--645.

\bibitem[{He et~al.(2021)He, Ravula, Kanagal, and
  Ainslie}]{he-etal-2021-realformer}
Ruining He, Anirudh Ravula, Bhargav Kanagal, and Joshua Ainslie. 2021.
\newblock {R}eal{F}ormer: Transformer likes residual attention.
\newblock In \emph{Findings of the Association for Computational Linguistics:
  ACL-IJCNLP 2021}, pages 929--943.

\bibitem[{Huang et~al.(2020)Huang, Perez, Ba, and Volkovs}]{pmlr-v119-huang20f}
Xiao~Shi Huang, Felipe Perez, Jimmy Ba, and Maksims Volkovs. 2020.
\newblock Improving transformer optimization through better initialization.
\newblock In \emph{Proceedings of the 37th International Conference on Machine
  Learning}, volume 119, pages 4475--4483.

\bibitem[{Ioffe and Szegedy(2015)}]{pmlr-v37-ioffe15}
Sergey Ioffe and Christian Szegedy. 2015.
\newblock Batch normalization: Accelerating deep network training by reducing
  internal covariate shift.
\newblock In \emph{Proceedings of the 32nd International Conference on Machine
  Learning (ICML)}, volume~37, pages 448--456.

\bibitem[{Klein et~al.(2017)Klein, Kim, Deng, Senellart, and
  Rush}]{klein-etal-2017-opennmt}
Guillaume Klein, Yoon Kim, Yuntian Deng, Jean Senellart, and Alexander Rush.
  2017.
\newblock {O}pen{NMT}: Open-source toolkit for neural machine translation.
\newblock In \emph{Proceedings of the 55th Annual Meeting of the Association
  for Computational Linguistics (ACL)}, pages 67--72.

\bibitem[{Kudo and Richardson(2018)}]{kudo-richardson-2018-sentencepiece}
Taku Kudo and John Richardson. 2018.
\newblock {S}entence{P}iece: A simple and language independent subword
  tokenizer and detokenizer for neural text processing.
\newblock In \emph{Proceedings of the 2018 Conference on Empirical Methods in
  Natural Language Processing (EMNLP)}, pages 66--71.

\bibitem[{Liu et~al.(2020)Liu, Liu, Gao, Chen, and
  Han}]{liu-etal-2020-understanding}
Liyuan Liu, Xiaodong Liu, Jianfeng Gao, Weizhu Chen, and Jiawei Han. 2020.
\newblock Understanding the difficulty of training transformers.
\newblock In \emph{Proceedings of the 2020 Conference on Empirical Methods in
  Natural Language Processing (EMNLP)}, pages 5747--5763.

\bibitem[{Merity et~al.(2017)Merity, Xiong, Bradbury, and
  Socher}]{DBLP:journals/corr/MerityXBS16}
Stephen Merity, Caiming Xiong, James Bradbury, and Richard Socher. 2017.
\newblock {Pointer Sentinel Mixture Models}.
\newblock In \emph{Proceedings of the 5th International Conference on Learning
  Representations (ICLR)}.

\bibitem[{Napoles et~al.(2012)Napoles, Gormley, and
  Van~Durme}]{napoles:2012:AG}
Courtney Napoles, Matthew Gormley, and Benjamin Van~Durme. 2012.
\newblock {Annotated Gigaword}.
\newblock In \emph{Proceedings of the Joint Workshop on Automatic Knowledge
  Base Construction and Web-scale Knowledge Extraction (AKBC-WEKEX)}, pages
  95--100.

\bibitem[{Ott et~al.(2019)Ott, Edunov, Baevski, Fan, Gross, Ng, Grangier, and
  Auli}]{ott-etal-2019-fairseq}
Myle Ott, Sergey Edunov, Alexei Baevski, Angela Fan, Sam Gross, Nathan Ng,
  David Grangier, and Michael Auli. 2019.
\newblock fairseq: A fast, extensible toolkit for sequence modeling.
\newblock In \emph{Proceedings of the 2019 Conference of the North {A}merican
  Chapter of the Association for Computational Linguistics (NAACL)}, pages
  48--53.

\bibitem[{Ott et~al.(2018)Ott, Edunov, Grangier, and
  Auli}]{ott-etal-2018-scaling}
Myle Ott, Sergey Edunov, David Grangier, and Michael Auli. 2018.
\newblock Scaling neural machine translation.
\newblock In \emph{Proceedings of the Third Conference on Machine Translation
  (WMT)}, pages 1--9.

\bibitem[{Panayotov et~al.(2015)Panayotov, Chen, Povey, and
  Khudanpur}]{7178964}
Vassil Panayotov, Guoguo Chen, Daniel Povey, and Sanjeev Khudanpur. 2015.
\newblock Librispeech: An asr corpus based on public domain audio books.
\newblock In \emph{2015 IEEE International Conference on Acoustics, Speech and
  Signal Processing (ICASSP)}, pages 5206--5210.

\bibitem[{Post(2018)}]{post-2018-call}
Matt Post. 2018.
\newblock A call for clarity in reporting {BLEU} scores.
\newblock In \emph{Proceedings of the Third Conference on Machine Translation
  (WMT)}, pages 186--191.

\bibitem[{Rush et~al.(2015)Rush, Chopra, and
  Weston}]{rush-chopra-weston:2015:EMNLP}
Alexander~M. Rush, Sumit Chopra, and Jason Weston. 2015.
\newblock {A Neural Attention Model for Abstractive Sentence Summarization}.
\newblock In \emph{Proceedings of the 2015 Conference on Empirical Methods in
  Natural Language Processing (EMNLP)}, pages 379--389.

\bibitem[{Schwartz et~al.(2019)Schwartz, Dodge, Smith, and
  Etzioni}]{DBLP:journals/corr/abs-1907-10597}
Roy Schwartz, Jesse Dodge, Noah~A. Smith, and Oren Etzioni. 2019.
\newblock Green {AI}.
\newblock \emph{CoRR}, abs/1907.10597.

\bibitem[{Sennrich et~al.(2016)Sennrich, Haddow, and
  Birch}]{sennrich-etal-2016-neural}
Rico Sennrich, Barry Haddow, and Alexandra Birch. 2016.
\newblock Neural machine translation of rare words with subword units.
\newblock In \emph{Proceedings of the 54th Annual Meeting of the Association
  for Computational Linguistics (ACL)}, pages 1715--1725.

\bibitem[{Srivastava et~al.(2015)Srivastava, Greff, and
  Schmidhuber}]{NIPS2015_215a71a1}
Rupesh~K Srivastava, Klaus Greff, and J\"{u}rgen Schmidhuber. 2015.
\newblock Training very deep networks.
\newblock In \emph{Advances in Neural Information Processing Systems 28
  (NIPS)}, pages 2377–--2385.

\bibitem[{Strubell et~al.(2019)Strubell, Ganesh, and
  McCallum}]{strubell-etal-2019-energy}
Emma Strubell, Ananya Ganesh, and Andrew McCallum. 2019.
\newblock Energy and policy considerations for deep learning in {NLP}.
\newblock In \emph{Proceedings of the 57th Annual Meeting of the Association
  for Computational Linguistics (ACL)}, pages 3645--3650.

\bibitem[{Takase and Kiyono(2021)}]{takase-kiyono-2021-rethinking}
Sho Takase and Shun Kiyono. 2021.
\newblock Rethinking perturbations in encoder-decoders for fast training.
\newblock In \emph{Proceedings of the 2021 Conference of the North American
  Chapter of the Association for Computational Linguistics: Human Language
  Technologies (NAACL-HLT)}, pages 5767--5780.

\bibitem[{Takase and Okazaki(2019)}]{takase-okazaki-2019-positional}
Sho Takase and Naoaki Okazaki. 2019.
\newblock Positional encoding to control output sequence length.
\newblock In \emph{Proceedings of the 2019 Conference of the North {A}merican
  Chapter of the Association for Computational Linguistics: Human Language
  Technologies (NAACL)}, pages 3999--4004.

\bibitem[{Vaswani et~al.(2018)Vaswani, Bengio, Brevdo, Chollet, Gomez, Gouws,
  Jones, Kaiser, Kalchbrenner, Parmar, Sepassi, Shazeer, and
  Uszkoreit}]{vaswani-etal-2018-tensor2tensor}
Ashish Vaswani, Samy Bengio, Eugene Brevdo, Francois Chollet, Aidan Gomez,
  Stephan Gouws, Llion Jones, {\L}ukasz Kaiser, Nal Kalchbrenner, Niki Parmar,
  Ryan Sepassi, Noam Shazeer, and Jakob Uszkoreit. 2018.
\newblock {T}ensor2{T}ensor for neural machine translation.
\newblock In \emph{Proceedings of the 13th Conference of the Association for
  Machine Translation in the {A}mericas}, pages 193--199.

\bibitem[{Vaswani et~al.(2017)Vaswani, Shazeer, Parmar, Uszkoreit, Jones,
  Gomez, Kaiser, and Polosukhin}]{NIPS2017_7181}
Ashish Vaswani, Noam Shazeer, Niki Parmar, Jakob Uszkoreit, Llion Jones,
  Aidan~N Gomez, \L~ukasz Kaiser, and Illia Polosukhin. 2017.
\newblock Attention is all you need.
\newblock In \emph{Advances in Neural Information Processing Systems 30
  (NIPS)}, pages 5998--6008.

\bibitem[{Wang et~al.(2020)Wang, Tang, Ma, Wu, Okhonko, and
  Pino}]{wang-etal-2020-fairseq}
Changhan Wang, Yun Tang, Xutai Ma, Anne Wu, Dmytro Okhonko, and Juan Pino.
  2020.
\newblock Fairseq {S}2{T}: Fast speech-to-text modeling with fairseq.
\newblock In \emph{Proceedings of the 1st Conference of the Asia-Pacific
  Chapter of the Association for Computational Linguistics and the 10th
  International Joint Conference on Natural Language Processing (AACL-IJCNLP)},
  pages 33--39.

\bibitem[{Wang et~al.(2022)Wang, Ma, Dong, Huang, Zhang, and
  Wei}]{https://doi.org/10.48550/arxiv.2203.00555}
Hongyu Wang, Shuming Ma, Li~Dong, Shaohan Huang, Dongdong Zhang, and Furu Wei.
  2022.
\newblock Deepnet: Scaling transformers to 1,000 layers.

\bibitem[{Wang et~al.(2019)Wang, Li, Xiao, Zhu, Li, Wong, and
  Chao}]{wang-etal-2019-learning-deep}
Qiang Wang, Bei Li, Tong Xiao, Jingbo Zhu, Changliang Li, Derek~F. Wong, and
  Lidia~S. Chao. 2019.
\newblock Learning deep transformer models for machine translation.
\newblock In \emph{Proceedings of the 57th Annual Meeting of the Association
  for Computational Linguistics (ACL)}, pages 1810--1822.

\bibitem[{Xiong et~al.(2020)Xiong, Yang, He, Zheng, Zheng, Xing, Zhang, Lan,
  Wang, and Liu}]{DBLP:conf/icml/XiongYHZZXZLWL20}
Ruibin Xiong, Yunchang Yang, Di~He, Kai Zheng, Shuxin Zheng, Chen Xing,
  Huishuai Zhang, Yanyan Lan, Liwei Wang, and Tie{-}Yan Liu. 2020.
\newblock On layer normalization in the transformer architecture.
\newblock In \emph{Proceedings of the 37th International Conference on Machine
  Learning (ICML)}, pages 10524--10533.

\bibitem[{Xu et~al.(2019)Xu, Sun, Zhang, Zhao, and Lin}]{NEURIPS2019_2f4fe03d}
Jingjing Xu, Xu~Sun, Zhiyuan Zhang, Guangxiang Zhao, and Junyang Lin. 2019.
\newblock Understanding and improving layer normalization.
\newblock In \emph{Advances in Neural Information Processing Systems
  (NeurIPS)}, volume~32.

\bibitem[{Zellers et~al.(2019)Zellers, Holtzman, Rashkin, Bisk, Farhadi,
  Roesner, and Choi}]{NEURIPS2019_3e9f0fc9}
Rowan Zellers, Ari Holtzman, Hannah Rashkin, Yonatan Bisk, Ali Farhadi,
  Franziska Roesner, and Yejin Choi. 2019.
\newblock Defending against neural fake news.
\newblock In \emph{Advances in Neural Information Processing Systems 32
  (NeurIPS)}, pages 9054--9065.

\bibitem[{Zhang et~al.(2019{\natexlab{a}})Zhang, Titov, and
  Sennrich}]{zhang-etal-2019-improving}
Biao Zhang, Ivan Titov, and Rico Sennrich. 2019{\natexlab{a}}.
\newblock Improving deep transformer with depth-scaled initialization and
  merged attention.
\newblock In \emph{Proceedings of the 2019 Conference on Empirical Methods in
  Natural Language Processing and the 9th International Joint Conference on
  Natural Language Processing (EMNLP-IJCNLP)}, pages 898--909.

\bibitem[{Zhang et~al.(2019{\natexlab{b}})Zhang, Dauphin, and
  Ma}]{conf/iclr/ZhangDM19}
Hongyi Zhang, Yann~N. Dauphin, and Tengyu Ma. 2019{\natexlab{b}}.
\newblock Fixup initialization: Residual learning without normalization.
\newblock In \emph{Proceedings of the 7th International Conference on Learning
  Representations (ICLR)}.

\end{thebibliography}

\clearpage

\appendix

\section{Details of Experimental Settings}
\label{sec:detail_setting}

\begin{table*}[!t]
  \centering
  \footnotesize
  \begin{tabular}{ l | c c c c} \hline
  Params & Machine Translation & Abstractive Summarization & Language Model & ASR \\ \hline
  Hidden dim size & 512 & 512 & 1024 & 512 \\
  FFN dim size & 2048 & 2048 & 4096 & 2048 \\
  Attention heads & 8 & 8 & 8 & 8 \\ \hline
  Learning rate & 0.001 & 0.001 & 0.001 & 0.001 \\
  Scheduler & inverse sqrt & inverse sqrt & inverse sqrt & inverse sqrt \\
  Adam \ $\beta$ & (0.9, 0.98) & (0.9, 0.98) & (0.9, 0.98) & (0.9, 0.98) \\
  Warmup updates & 4K & 4K & 2K & 4K \\
  Max updates & 50K & 50K & 50K & 150K \\
  Max tokens / GPU & 3584 & 3584 & 1024 & 40K \\ \hline
  \end{tabular}
  \caption{Hyper-parameters used in our experiments.\label{tab:hyp_params}}
\end{table*}

\begin{table*}[!t]
  \centering
  \footnotesize
  \begin{tabular}{ l | c c c c c c c c} \hline
  & \multicolumn{2}{c}{Machine Translation} & \multicolumn{2}{c}{Abstractive Summarization} & \multicolumn{2}{c}{Language Model} & \multicolumn{2}{c}{ASR} \\
  & 6L-6L & 18L-18L & 6L-6L & 18L-18L & 6L & 16L & 6L-6L & 12L-6L \\ \hline
  \#GPU & 128 & 128 & 64 & 144 & 128 & 192 & 32 & 32 \\
  Time (hour) & 5 & 13 & 4 & 17 & 4 & 7 & 22 & 34 \\ \hline
  \end{tabular}
  \caption{The number of GPUs and computational time used to construct one model in our experiments.\label{tab:comp_resource}}
\end{table*}

\subsection{Hyper-parameters}
\label{sec:hyp_params}
As described in Section \ref{sec:experiments}, our hyper-parameters follow those used in previous studies.
Table \ref{tab:hyp_params} shows hyper-parameters used for each experiment.
For fair comparisons, we used the same hyper-parameters for all methods except T-Fixup.
For T-Fixup, we used hyper-parameters reported in \citet{pmlr-v119-huang20f} to prevent divergence.

\subsection{Computational Resources}
\label{sec:comp_res}
We mainly used NVIDIA Tesla P100 GPUs for most of our experiments.
Table \ref{tab:comp_resource} shows the number of GPUs and the computational time used to construct one model in our experiments.
For the 100L-100L configuration, described in Section \ref{sec:exp_mt}, we used 24 Tesla V100 GPUs and spent approximately 120 hours to train one model.

\section{Supplementary of Gradient Norms of Each Location}
\label{sec:appendix_grad}

\begin{figure*}[!t]
  \centering 
  \includegraphics[width=14cm]{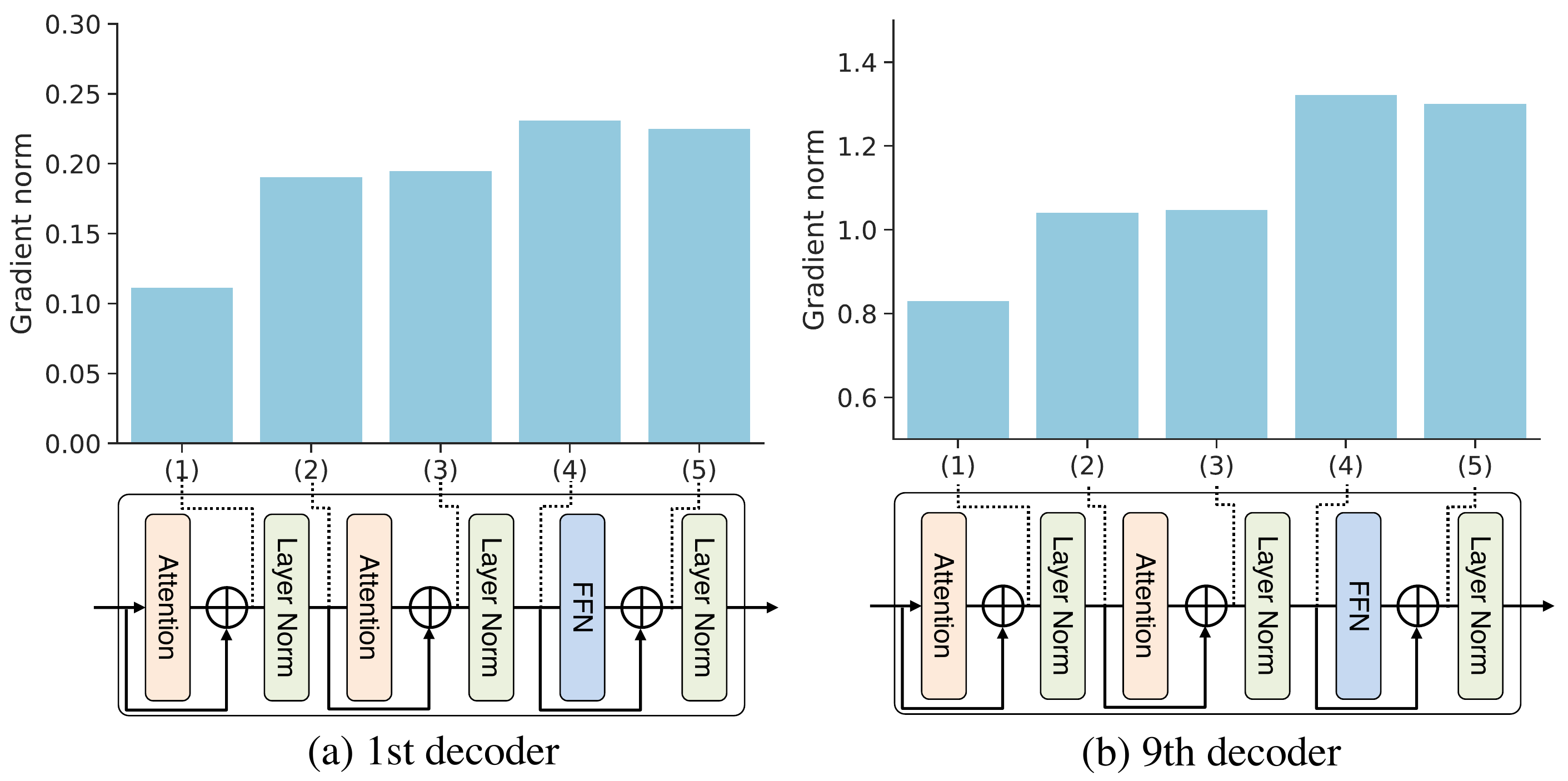}
   \caption{Gradient norms of each part in the (a) 1st decoder and (b) 9th decoder of the 18L-18L Post-LN Transformer encoder-decoder on WMT English-to-German translation training data.}
   \label{fig:grad4eachpart_additional}
\end{figure*}

For gradient norms of each part in a layer, we check 1st and 9th decoders in addition to the 18th decoder for the 18L-18L Post-LN Transformer encoder-decoder as shown in Figure \ref{fig:grad4eachpart}.
Figure \ref{fig:grad4eachpart_additional} shows the gradient norms of each part.
This figure shows that the gradient norms decrease drastically through layer normalizations in the same manner as they do in the 18th decoder (Figure \ref{fig:grad4eachpart}).
Therefore, the vanishing gradient problem in Post-LN Transformers is probably caused by layer normalizations.

\section{Details of the 100L-100L Configuration}
\label{sec:stack100layer}

\subsection{Regularizations during the Training}
\begin{figure}[!t]
  \centering 
    \includegraphics[width=7cm]{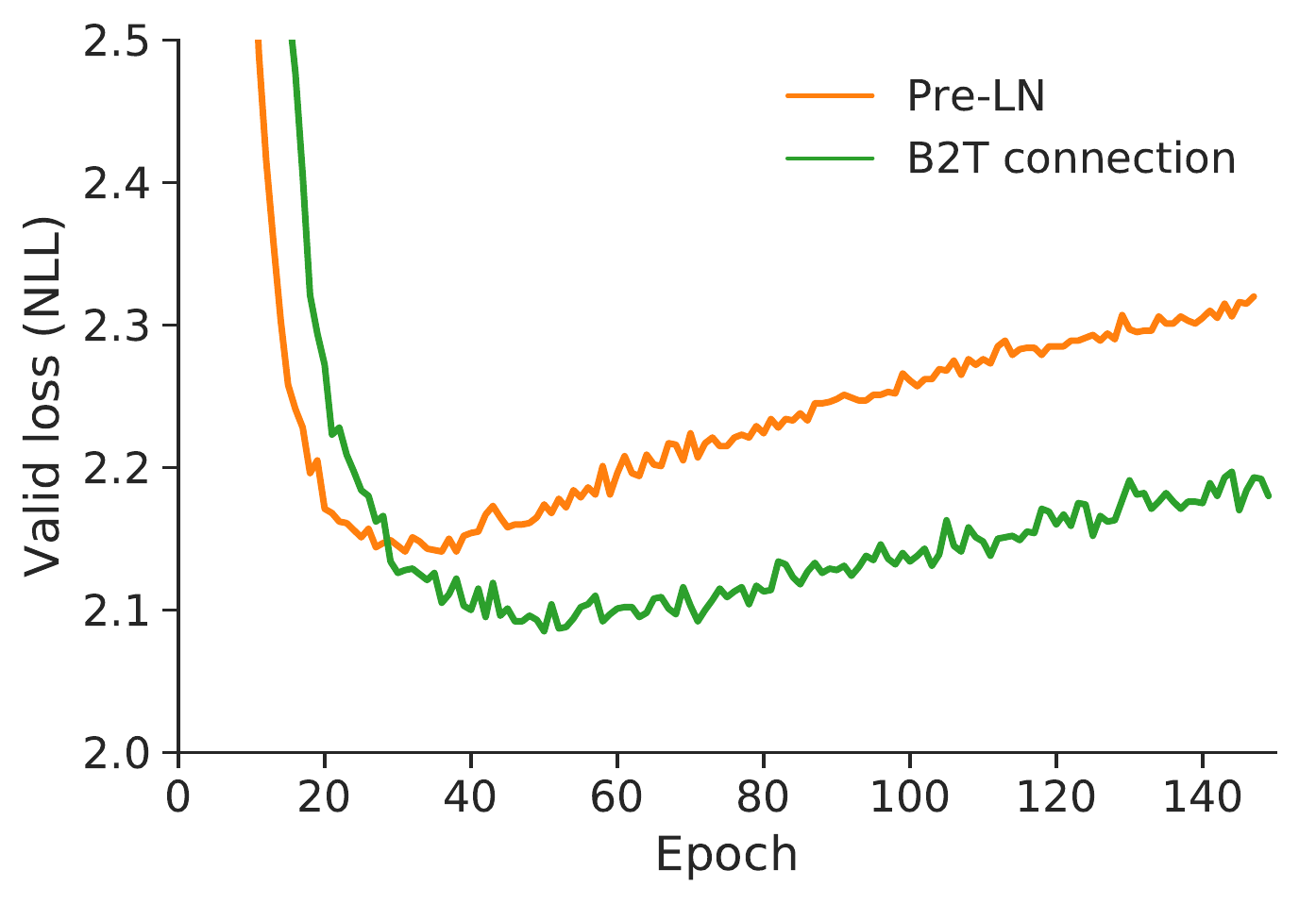}
   \caption{Negative Log-Likelihood (NLL) values of Pre-LN and our proposed B2T connection on validation data (newstest2013) in 36L-36L. We used the same hyper-parameters as those used in 6L-6L and 18L-18L.\label{fig:training_curve_36l}}
\end{figure}

\begin{figure}[!t]
  \centering 
    \includegraphics[width=7cm]{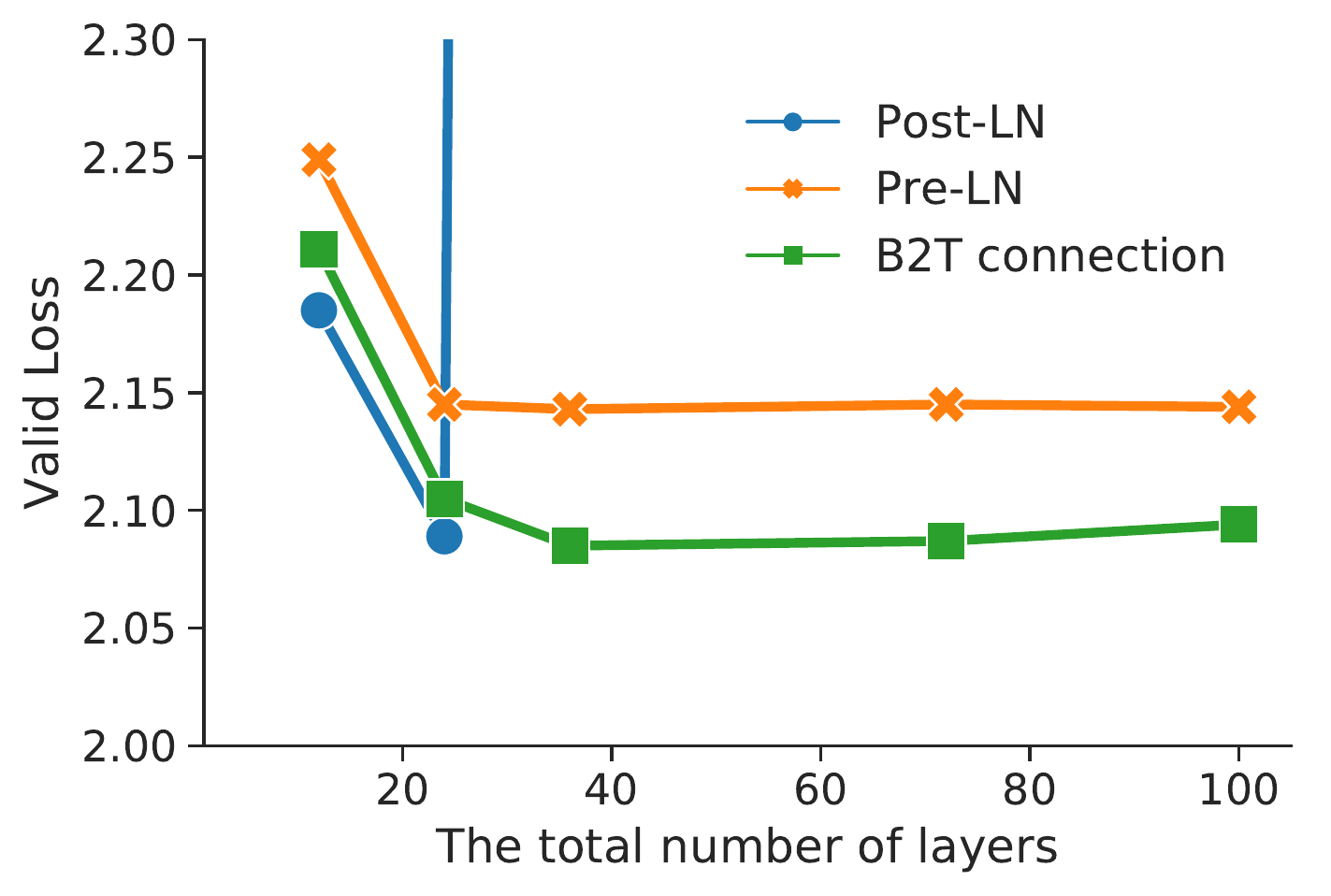}
   \caption{The best Negative Log-Likelihood (NLL) values on validation data (newstest2013) when the total number of layers is varied. The total number of layers is divided equally between the encoder and decoder.}
   \label{fig:valid_loss_each_layersize}
\end{figure}

As reported in Section \ref{tab:exp_mt}, we constructed 100L-100L Transformers with widely-used WMT English-to-German dataset.
In the preliminary experiments, we found that regularization is the key to preventing overfitting and achieving high performance in this situation.
Figure \ref{fig:training_curve_36l} shows the NLL values of Pre-LN and B2T connection on validation data in the 36L-36L configuration when we used the same hyper-parameters as those used in 6L-6L and 18L-18L configurations.
As this figure shows, the NLL values began to increase from the middle of training, and thus the overfitting occurred.
In addition, the use of the same hyper-parameters as 6L-6L and 18L-18L makes it difficult to improve the performance of deeper configurations.
Figure \ref{fig:valid_loss_each_layersize} shows the best NLL values on validation data when we varied the number of layers: 6L-6L, 12L-12L, 18L-18L, 36L-36L, and 50L-50L\footnote{The horizontal axis of Figure \ref{fig:valid_loss_each_layersize} represents the total number of layers, which are divided equally between the encoder and decoder. For example, 100 on the horizontal axis represents 50L-50L Transformers.}.
This figure indicates that adding more layers to the 18L-18L configuration did not improve the performance.

To prevent overfitting during the training of 100L-100L Transformers, we increased the dropout rate from $0.3$ to $0.5$.
In addition, we used word dropout, as described in \newcite{takase-kiyono-2021-rethinking}.
We set the word dropout rate to $0.1$ for the encoder and decoder.
We multiplied the initial parameter values, except those for embeddings, by $0.1$.
We set the gradient clipping to $0.1$.
Finally, we decreased the number of updates from 50K to 25K.
These regularization techniques prevented overfitting and achieved better performance than 18L-18L, as described in Section \ref{sec:exp_mt}.

\subsection{Comparison with DeepNet}

\begin{figure}[!t]
  \centering 
    \includegraphics[width=7cm]{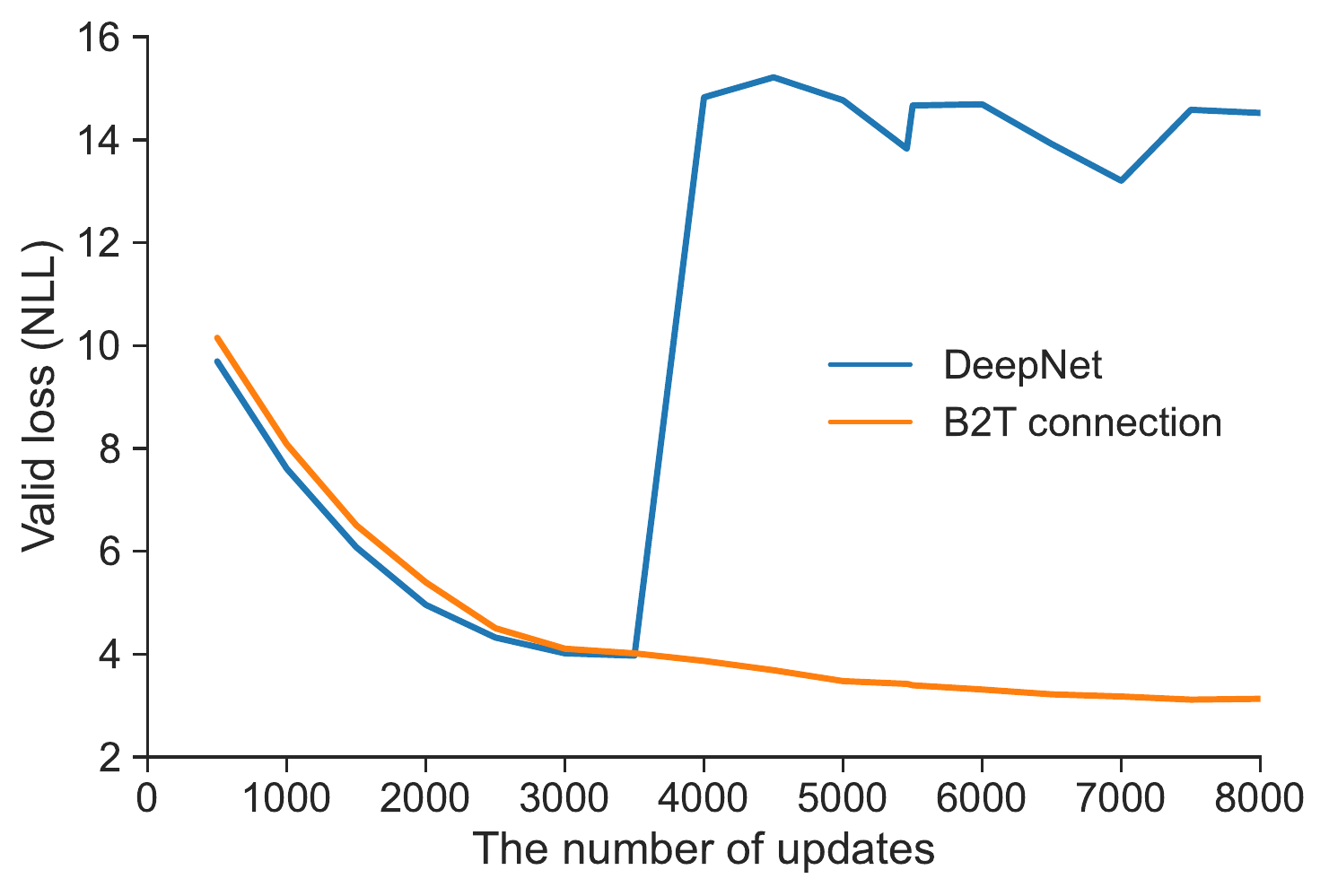}
   \caption{Negative Log-Likelihood (NLL) values of DeepNet and our proposed B2T connection on validation data (newstest2013) in 100L-100L with a small batch size.}
   \label{fig:training_curve_100l}
\end{figure}

As described in Section \ref{sec:related_work}, various studies have attempted to stabilize the training of deep Transformers.
Each study indicated the effectiveness of their proposed method empirically, and some have provided theoretical justifications.
However, \newcite{https://doi.org/10.48550/arxiv.2203.00555} demonstrated that the training of previous methods except DeepNet failed in a much deeper configuration than normally used, i.e., 100L-100L.
Then, can we conclude that DeepNet is a silver bullet for deep Transformers?
It is difficult to reach this conclusion because the training of DeepNet also fails in some configurations.
For example, when we train deep Transformers, we might decrease the batch size because the trainable parameters occupy most of the GPU memories.
When we tried this, the NLL value of DeepNet on validation data diverged, as shown in Figure \ref{fig:training_curve_100l}.
In other words, the training of DeepNet failed.
In contrast, the training of our B2T connection succeeded in this situation.
This result implies that there are problems in the training of deep Transformers that have not been solved in previous studies.
Therefore, we believe that we should continue to add the empirical findings about new techniques, including B2T connection, to those of previous studies.

\section{B2T Connection without Layer Normalization}
In addition to B2T connection, we also consider a further modification to prevent the vanishing gradient problem.
Because layer normalizations decrease gradients drastically, as described in Section \ref{sec:gradient}, removing layer normalizations may provide stable gradients during back-propagation.
However, the values in the forward pass increase exponentially if layer normalizations are removed.
Therefore, we introduce weights that prevent the explosive increase in the forward pass while mitigating the decreasing gradients in back-propagation, as an alternative to the layer normalization.
To use this alternative, we replace Equation (\ref{eq:b2t}) with the following equation:
\begin{align}
\alpha x_{inp} + \beta \left( x_{ffn} + \mathrm{FFN}(x_{ffn}) \right).\label{eq:w/o_ln}
\end{align}
Through several experiments\footnote{We could instead tune $\alpha$ and $\beta$ to improve performance on each task but here we define values that are useful for various tasks.}, we found that the following values of $\alpha$ and $\beta$ are suitable:
\begin{align}
\alpha &= \mathrm{min}\left(\frac{N}{12}, N^{-0.15}  \right) ,\\
\beta &= d^{-0.2},
\end{align}
where $N$ is the number of layers and $d$ is the dimension of the input vectors $x_{inp}$.
For example, $N$ is set to 12 and 6 in the encoder and decoder, respectively, in the 12L-6L configuration.
Therefore, as the number of layers increases, the value of $\alpha$ increases while $N$ remains small (until $N=9$), and then $\alpha$ starts to decrease.
In short, $\alpha$ prevents an explosive increase in the forward pass when we stack many layers.
$\beta$ decreases as the dimension $d$ increases, and thus it prevents an explosive increase when a large dimension is used.
By using Equation (\ref{eq:w/o_ln}), we can remove all layer normalizations in internal layers.
This solves the vanishing gradient problem caused by layer normalizations.

\begin{table*}[!t]
  \centering{}
  \footnotesize
  \begin{tabular}{ l | c c c c c c c | c} \hline
  Method & 2010 & 2011 & 2012 & 2013 & 2014 & 2015 & 2016 & Average \\ \hline \hline
  \multicolumn{9}{c}{Enc-Dec: 6L-6L} \\ \hline \hline
  B2T connection & 24.12 & 21.93 & \textbf{22.29} & \textbf{26.31} & 26.84 & 29.48 & \textbf{34.73} & \textbf{26.53} \\
  + w/o LN & \textbf{24.17} & \textbf{22.07} & 22.24 & 25.83 & \textbf{26.96} & \textbf{29.70} & 34.42 & 26.48 \\ \hline \hline
  \multicolumn{9}{c}{Enc-Dec: 18L-18L} \\ \hline \hline
  B2T connection & \textbf{24.75} & \textbf{22.88} & \textbf{23.09} & \textbf{27.12} & \textbf{28.82} & \textbf{30.99} & \textbf{33.64} & \textbf{27.33} \\
  + w/o LN & 24.47 & 22.37 & 22.58 & 27.04 & 28.34 & 30.49 & \textbf{34.38} & 27.10 \\ \hline
  \end{tabular}
  \caption{BLEU scores of our modifications on WMT newstest2010-2016 and their averages.\label{tab:exp_mt_woln}}
\end{table*}

\begin{table}[!t]
  \centering{}
  \footnotesize
  \begin{tabular}{ l | c c c } \hline
  Method & R-1 & R-2 & R-L \\ \hline \hline
  \multicolumn{4}{c}{Enc-Dec: 6L-6L} \\ \hline \hline
  B2T connection & 38.43 & 19.37 & 35.72 \\
  + w/o LN & \textbf{38.63} & \textbf{19.75} & \textbf{35.77} \\ \hline \hline
  \multicolumn{4}{c}{Enc-Dec: 18L-18L} \\ \hline \hline
  B2T connection & \textbf{39.61} & \textbf{20.28} & \textbf{36.66} \\
  + w/o LN & 39.29 & 20.01 & 36.48 \\ \hline
  \end{tabular}
  \caption{F1 based ROUGE scores of our modifications on headline generation.\label{tab:exp_summarization_woln}}
\end{table}

Tables \ref{tab:exp_mt_woln} and \ref{tab:exp_summarization_woln} shows the results of B2T connection without layer normalizations (``w/o LN'') on the machine translation and summarization tasks.
These results indicate that B2T connection without layer normalizations achieved scores comparable to those of B2T connection with layer normalizations.
However, because the results of B2T connection without layer normalizations are slightly worse than those with layer normalizations, we recommend the use of B2T connection with layer normalizations.

\end{document}